\documentclass[10pt,twocolumn,letterpaper]{article}

\usepackage{cvpr}
\usepackage{times}
\usepackage{epsfig}
\usepackage{graphicx}
\usepackage{amsmath}
\usepackage{amssymb}

\usepackage{color}

\usepackage{hyperref}
\usepackage{graphicx}
\usepackage[export]{adjustbox}

\usepackage{amsfonts}
\usepackage{booktabs}
\usepackage{siunitx}
\usepackage{pifont}
\usepackage{subcaption}
\usepackage{diagbox}
\usepackage{multirow}
\usepackage{pifont}

\cvprfinalcopy 

\def\cvprPaperID{3264} 

\ifcvprfinal\fi
\begin{document}






\title{ARVo: Learning All-Range Volumetric Correspondence for Video Deblurring}

\author{Dongxu Li$^{*,1,2}$, Chenchen Xu$^{*,1,2}$, Kaihao Zhang$^{*,\dagger,1}$, \\Xin Yu$^{3}$, Yiran Zhong$^{1}$, Wenqi Ren$^{4}$, Hanna Suominen$^{1,5}$, Hongdong Li$^{1}$\\
$^{1}$ANU, $^{2}$DATA61-CSIRO, $^{3}$UTS, $^{4}$IIE-CAS, $^{5}$University of Turku\\
}


\newcommand{\argmin}{\operatornamewithlimits{argmin}}
\newcommand{\argmax}{\operatornamewithlimits{argmax}}

\twocolumn[{
\begin{@twocolumnfalse}
\maketitle
\begin{center}
\setlength\tabcolsep{1pt}
\vspace{-0.55cm}\begin{tabular}{cccccc}
\includegraphics[width=0.16\textwidth]{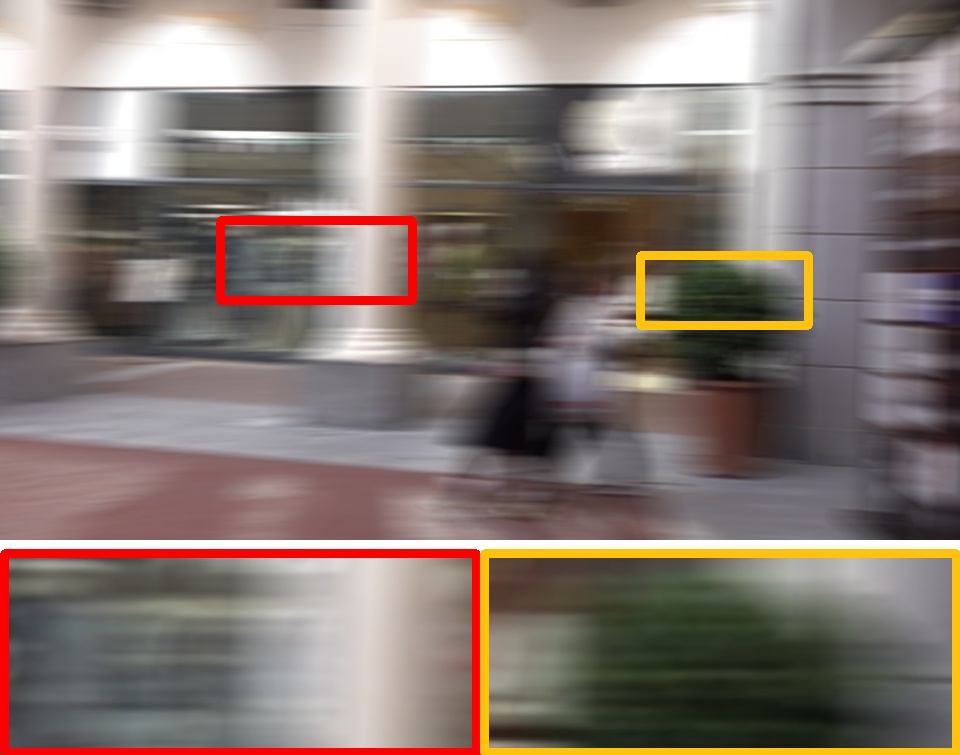}&
\includegraphics[width=0.16\textwidth]{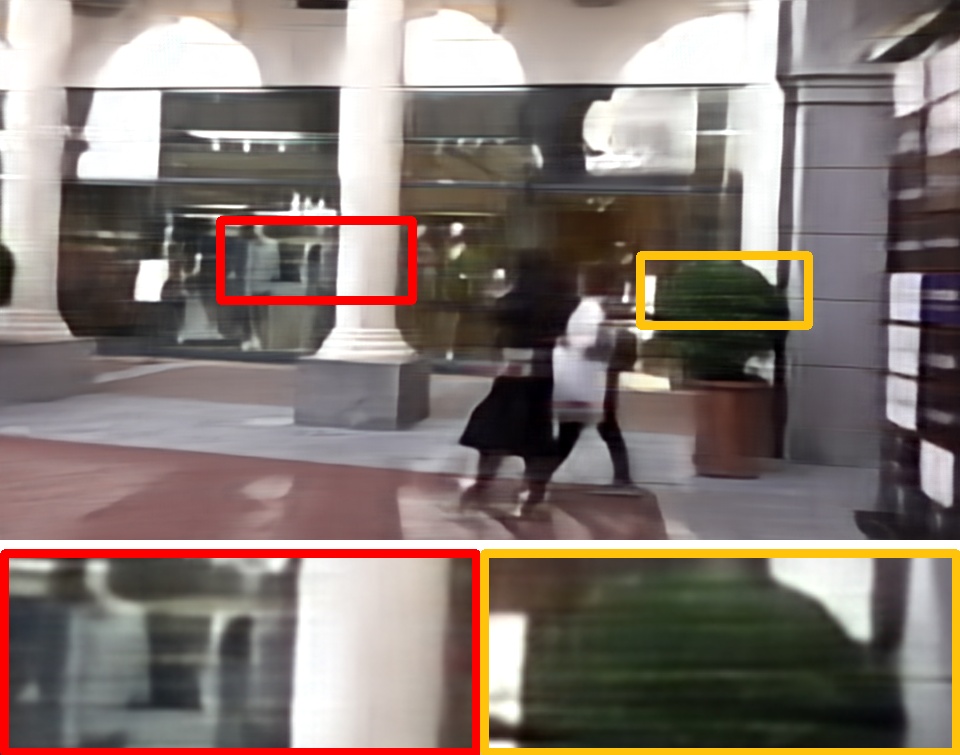}&
\includegraphics[width=0.16\textwidth]{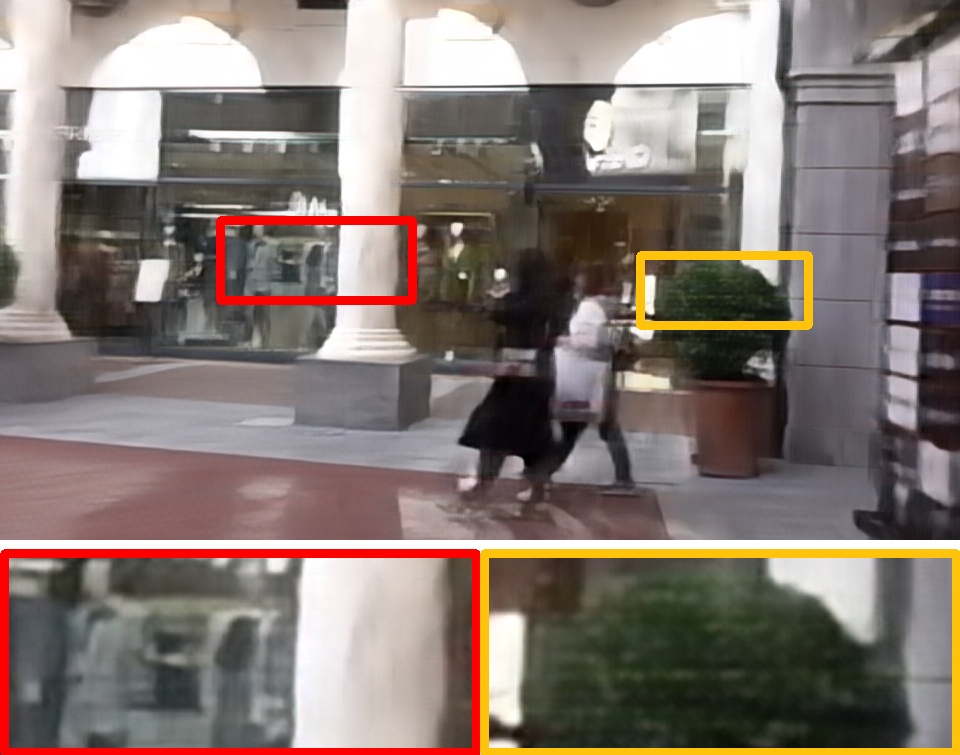}&
\includegraphics[width=0.16\textwidth]{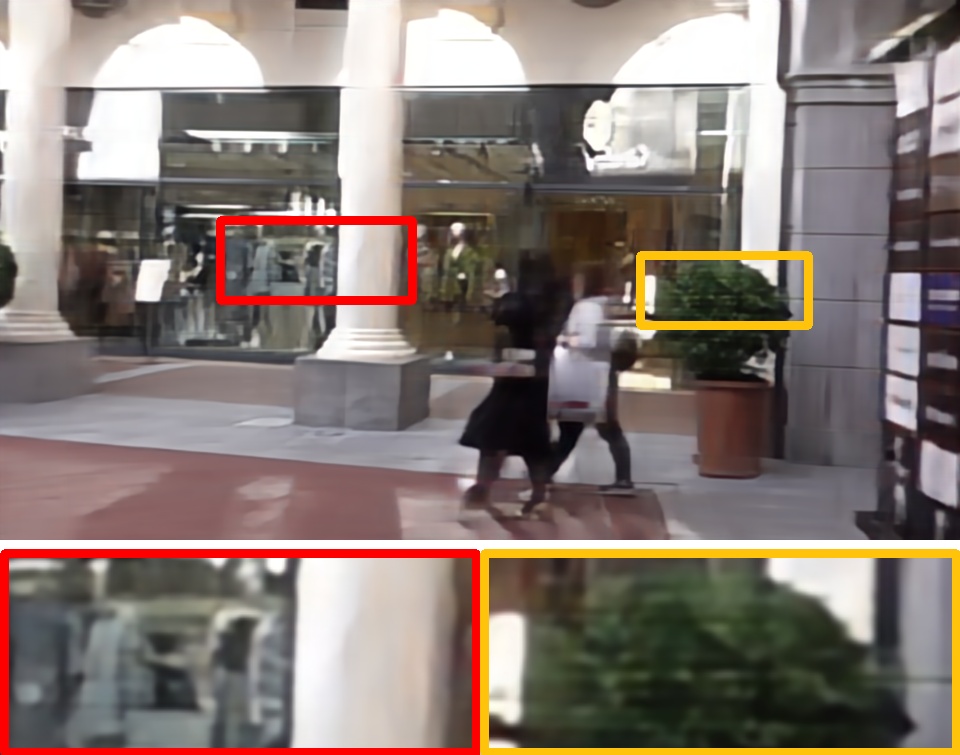}&
\includegraphics[width=0.16\textwidth]{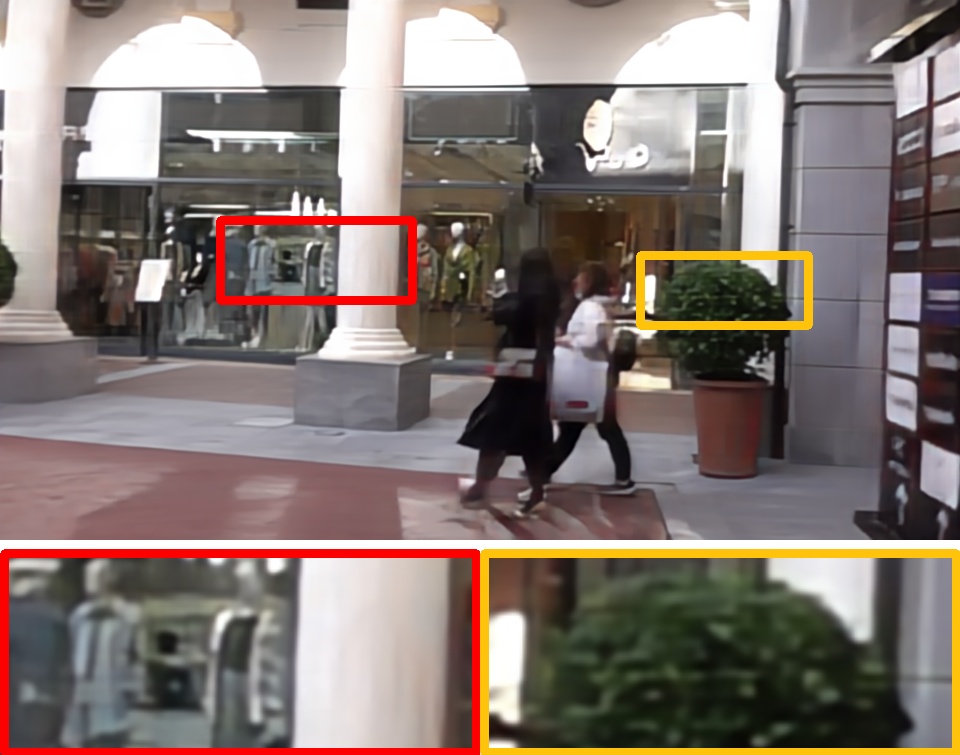}&
\includegraphics[width=0.16\textwidth]{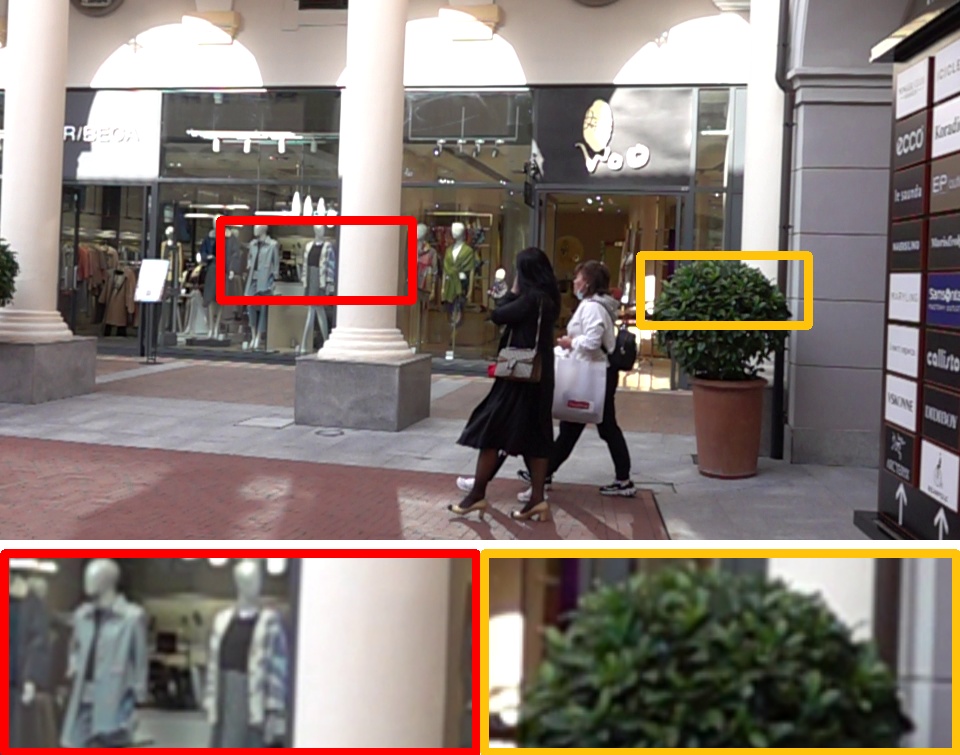}\\
\includegraphics[width=0.16\textwidth]{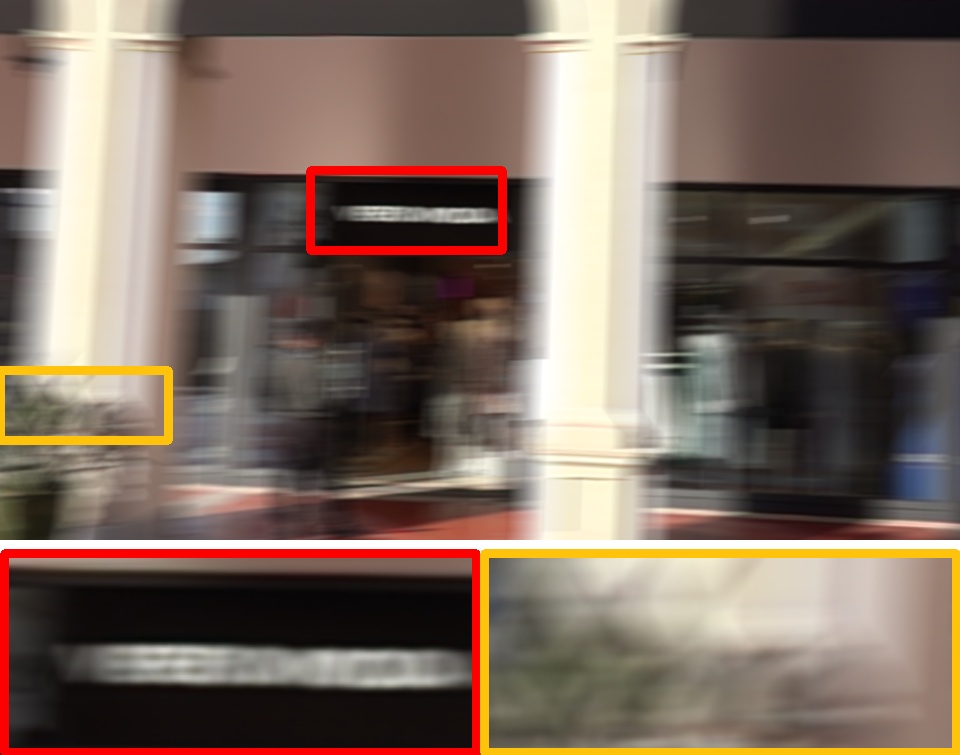} & 
\includegraphics[width=0.16\textwidth]{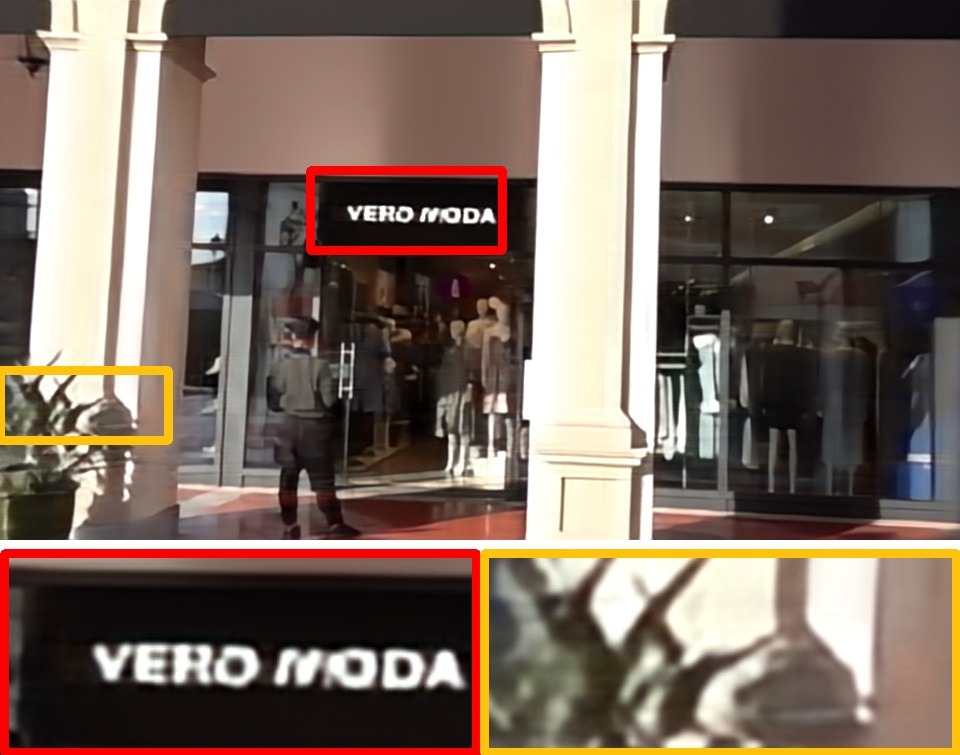} &
\includegraphics[width=0.16\textwidth]{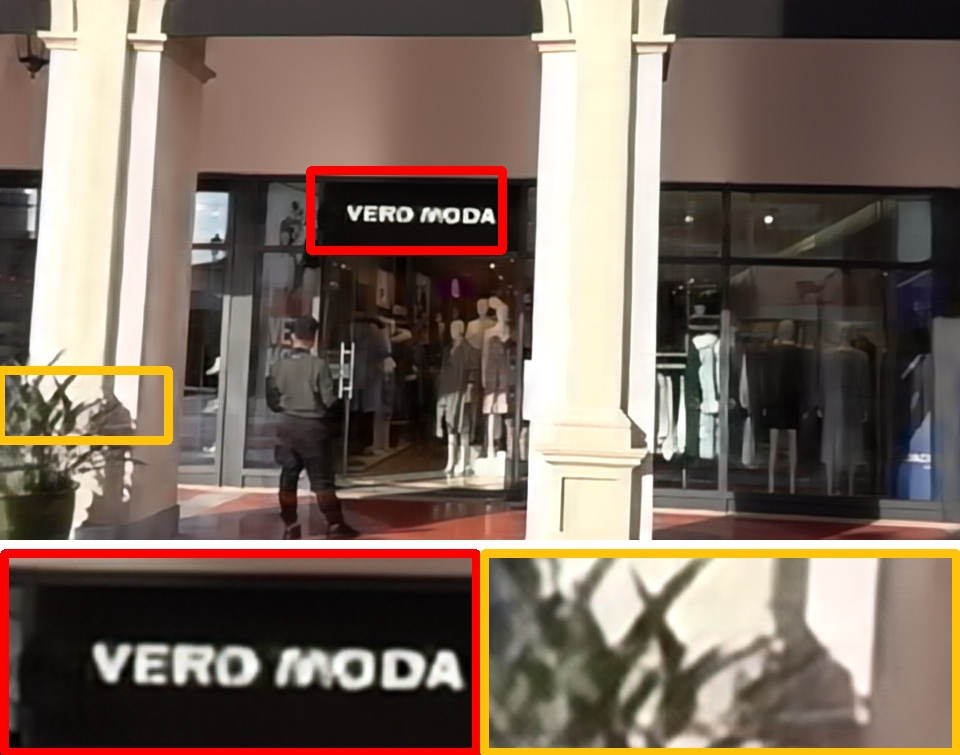} &
\includegraphics[width=0.16\textwidth]{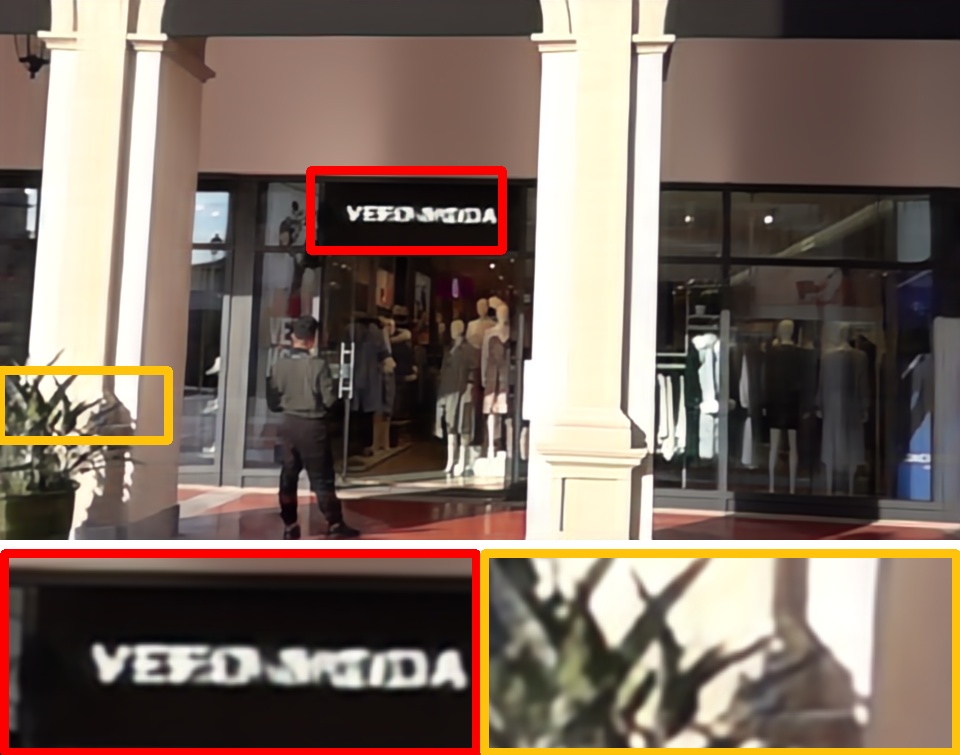}& 
\includegraphics[width=0.16\textwidth]{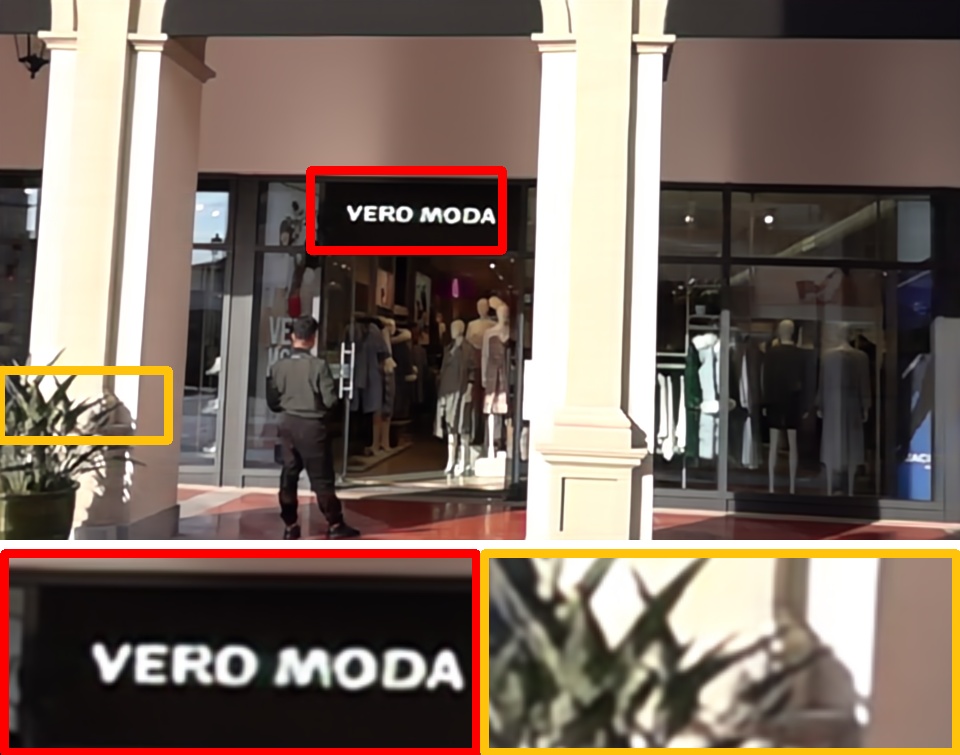}&
\includegraphics[width=0.16\textwidth]{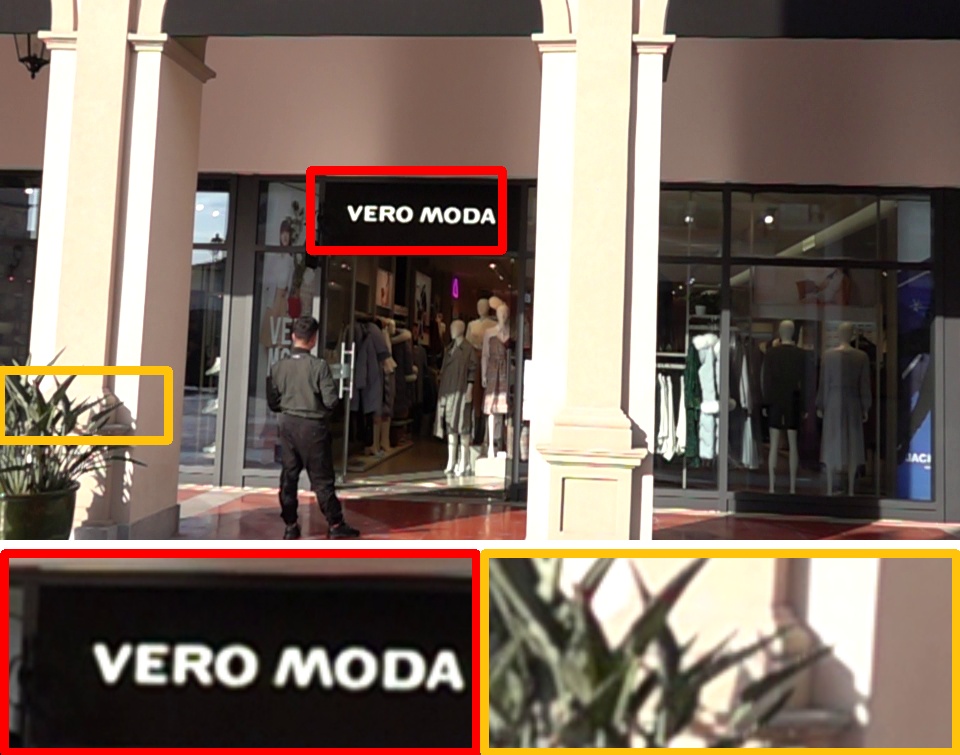}\\
(a) Input & (b) Su~\etal~\cite{su2017deep} & (c) STFAN~\cite{zhou2019spatio} & (d) TSP~\cite{pan2020cascaded} & (e) ARVo (Ours) & (f) Ground-truth

\end{tabular}
\end{center}
\vspace{-0.5cm}
\captionof{figure}{Qualitative comparisons of different models on the newly introduced HFR-DVD dataset. The dataset features sharp frames captured with a high frame-rate at 1,000 fps, providing more realistic motion blurs and less artefacts. Our model matches pixel-pairs across all the spatial range in the feature space, thus having a better capacity to build correspondence with large pixel displacements. This makes our model more favorable to handle blurs, especially due to fast motions.}
\vspace{1cm}
\label{fig:hfr}\end{@twocolumnfalse}
}]

\begin{abstract}\let\thefootnote\relax\footnotetext{$*$ Authors contributed equally. 

 ~~~$\dagger$ Corresponding author: \texttt{kaihao.zhang@anu.edu.au}.}
  Video deblurring models exploit consecutive frames to remove blurs from camera shakes and object motions. In order to utilize neighboring sharp patches, typical methods rely mainly on homography or optical flows to spatially align neighboring blurry frames. However, such explicit approaches are less effective in the presence of fast motions with large pixel displacements.
  In this work, we propose a novel implicit method to learn spatial correspondence among blurry frames in the feature space. 
  To construct distant pixel correspondences, our model builds a correlation volume pyramid among all the pixel-pairs between neighboring frames.
  %
  To enhance the features of the reference frame, we design a correlative aggregation module that maximizes the pixel-pair correlations with its neighbors based on the volume pyramid.
  %
  %
  Finally, we feed the aggregated features into a reconstruction module to obtain the restored frame. 
  %
  %
  %
  We design a generative adversarial paradigm to optimize the model progressively.
  Our proposed method is evaluated on the widely-adopted DVD dataset, along with a newly collected High-Frame-Rate (1000 fps) Dataset for Video Deblurring (HFR-DVD).
  Quantitative and qualitative experiments show that our model performs favorably on both datasets against previous state-of-the-art methods, confirming the benefit of modeling all-range spatial correspondence for video deblurring.
  
\end{abstract}
\section{Introduction}
Hand-held devices are popular in capturing videos of dynamic scenes, where prevalent high-speed object movements and abrupt camera shakes lead to undesirable blurs in videos.
To this end, the task of video deblurring aims to improve the video quality by restoring sharp frames from blurry video sequences and benefits a wide range of vision applications~\cite{li2020word,maxexp,hosvd,li2020transferring,li2020tspnet,li2020content,Zhang_2020_ACCV}.


Video deblurring methods remove blurs by taking advantage of sharper scene patches from neighboring frames~\cite{matsushita2006full,cho2012video}. 
However, since neighboring frames are usually not spatially aligned with the reference frame due to motions,  it is non-trivial to construct visual correspondence between frames.
Existing works mainly tackle the misalignment between frames using homography or optical flows.
%
Some previous works~\cite{su2017deep,hyun2015generalized,gong2017motion,sun2015learning,xiang2020deep,pan2020cascaded} first estimate the relative motions using optical flows, then warp neighboring frames to the reference frame.
However, in occurrence of fast object motions with large displacements, the constraint on the velocity smoothness may not always hold. As a result, estimating optical flow from blurry images remains a challenging research problem by itself~\cite{portz2012optical,tang2017depth}.
In addition, these methods are also notoriously less effective in the presence of occlusions and severe depth variations.
%
%
Recent works~\cite{jo2018deep,zhou2019spatio} use dynamic filters to restore videos in the feature domain. However, their ability to address large pixel displacement is restricted by the local receptive fields. 
Therefore, it remains a challenge to design a flexible yet effective visual correspondence method with abrupt motions.
%


To tackle this problem, as a complement to the existing pixel based warping methods, we propose an \emph{implicit} approach that estimates the pairwise image correspondence in the feature space.
%
%
%
%
In particular, we first extract visual features from a sequence of warped consecutive frames.
Then, we pair up the reference frame and its neighboring frame, and compute their pixel-wise correlations by matching their features in the embedding space.
Specifically, in order to account for distant pixel correspondence, we compute correlations for pixel pairs in all the spatial range.
In this way, we enable the model to better capture the visual dependencies at various lengths, thus being more effective in handling blurs to fast motions.

To further increase the receptive fields while maintaining the fine-grained visual details, we propose to build a pyramid of correlation volumes in order to achieve feature matching at different spatial scales.
Particularly, rather than subsampling all the frames to the same scale as in typical feature pyramid modules~\cite{he2015spatial}, we keep the reference frame feature maps while subsampling only those of neighboring frames. By maintaining the spatial scale of the reference frame, we keep high resolution information, allowing the model to learn correspondence taking into account the fine-grained visual details.

We optimize our model progressively in a generative adversarial paradigm with a new temporal consistency loss.
The training proceeds in stages, where later stages take as input the restored frames from the previous stage.
In this way, we ease the optimization by allowing the model to restore details gradually.
We propose an adversarial loss to encourage the temporal consistency between the restored frames.
Distinct from the previous adversarial video deblurring method~\cite{zhang2018adversarial} that takes as input a single restored frame, we use a discriminator to distinguish between restored frame sequence and the ground-truth sharp frame sequence.
We show the two training strategies combined improve the restoration quality quantitatively and visually.

To validate the effectiveness of the proposed method, we first conduct evaluations on the widely-adopted DVD dataset~\cite{su2017deep}.
In addition, since the DVD dataset uses automatic frame interpolation to increase the frame rate, it exhibits artefacts in the synthetic blurry frames.
To this end, we contribute a new large-scale dataset for video deblurring research, called \emph{HFR-DVD}. The HFR-DVD dataset is captured using a high-speed camera in 1,000 fps, featuring sharper frames and more realistic motion blurs.
Our experiments show that the proposed feature correlation methods help to improve the restoration on both datasets.

Our contributions are summarized as follows:
\begin{itemize}
  \setlength\itemsep{-0.2em}
    \item we introduce a novel video deblurring method by constructing spatial correspondence between pixel pairs in the feature space. To account for distant pixel displacements, we match pixel pairs in all the spatial range between the reference frame and neighboring frames; 
    \item we propose a correlative aggregation module to enhance the reference frame feature based on the correlation volume pyramid;
    \item in order to encourage the temporal consistency in the restored frames, we develop a adversarial loss for optimizing the model;
    \item we benchmark existing video deblurring methods on a new large-scale high-frame-rate dataset HFR-DVD with sharper frames and more realistic blurs;
    \item our video deblurring model ARVo outperforms previous methods quantitatively and qualitatively on the DVD and HFR-DVD datasets, establishing a new state-of-the-art for video deblurring. 
\end{itemize}






%

%
\section{Related Works}
We approach the video deblurring task by constructing dense correspondence between features.
Our work is thus broadly related to image deblurring
video deblurring
, and dense correspondence matching.
%
\begin{figure*}[ht]
\centering
  \includegraphics[width=\textwidth]{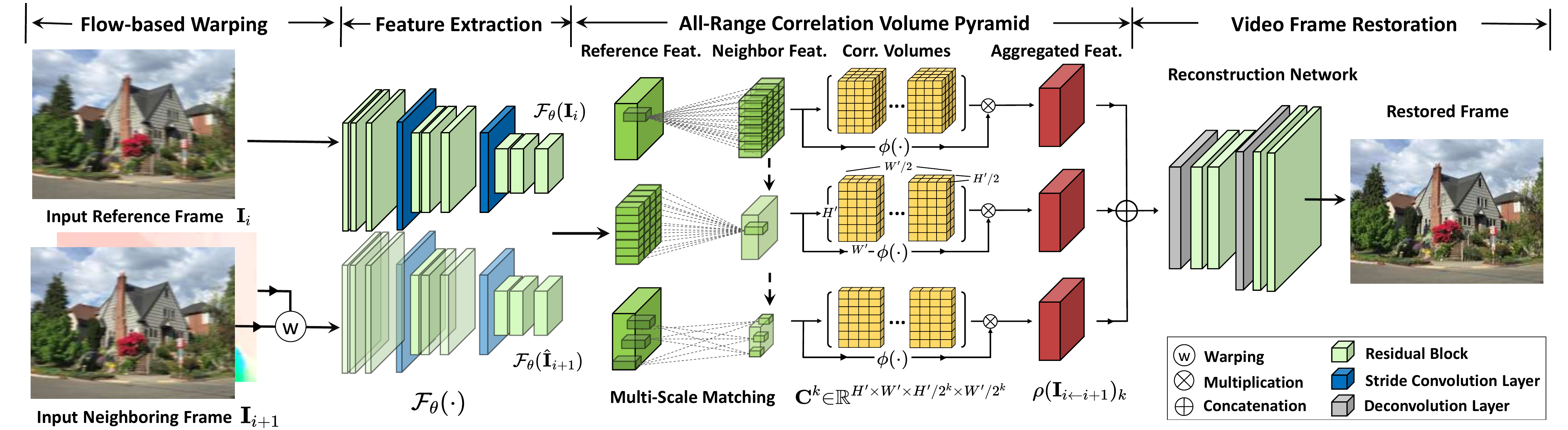}
\caption{Overview of the proposed video deblurring pipeline. Our method first warps the neighboring frame to the reference frame for feature extraction. Then it computes all-range volumetric correspondence between the reference and neighboring frames as a correlation volume pyramid.
Next, it relies on a correlative aggregation module to enhance the reference frame feature maps. Finally, we obtain the restored frame from a reconstruction network. Note that we only display one neighboring frame for illustration purposes, while our model in facts computes one intra-frame and two inter-frame  correlation volumes.}
\label{fig:arch}
\end{figure*}
~\label{fig:arch}

\noindent{\textbf{Image deblurring.}}~~
%
Early works on image deblurring assume a uniform blur kernel and mainly focus on designing effective image priors~\cite{shan2008high,krishnan2009fast,pan2016blind,yu2014efficient} to regularize the deblurring process into a well-posed problem. 
Recent deep learning models rely on paired sharp and blurry images for supervised training. 
In particular, the methods~\cite{sun2015learning,gong2017motion} learn to model the non-uniform blur kernels.  
Tao~\etal~\cite{tao2018scale} design a recurrent structure to iteratively refine the deblurring results.
Zhang~\etal~\cite{zhang2018dynamic} propose a spatially invariant recurrent network to remove blurs in the feature space. 
Gao~\etal~\cite{gao2019dynamic} propose a nested skip connection structure with selective parameter sharing.
These models show promising deblurring results on the synthetic blurry images.
The work~\cite{zhang2020deblurring} attempts to model the realistic blurs from real-life blurred images.
Most learning-based works are trained with the typical reconstruction loss, \eg~$L_1$ or $L_2$ losses, while~\cite{kupyn2018deblurgan} uses an additional adversarial loss.
%

\noindent{\textbf{Video deblurring.}}~~Early works on video deblurring mostly attempt to restore images by estimating blur kernels~\cite{li2010generating,zhang2013multi,zhang2014multi,hyun2015generalized}. 
Recently, Su~\etal~\cite{su2017deep} align neighboring frames using homography and optical flows and use a CNN to restore frames. 
Kim~\etal~\cite{hyun2017online} employ a recurrent neural network architecture and propose a dynamic temporal blending technique to enforce temporal consistency among frames.
Instead of explicit alignment, Zhou~\etal~\cite{zhou2019spatio} use dynamic filters to adaptively align frame features. However, their method is restricted by the limited local receptive field and is less effective for fast motions. 
The works~\cite{xiang2020deep,pan2020cascaded} estimate sharpness as a prior and utilize sharp images as exemplars to aid the learning of deblurring networks. 
Zhang~\etal~\cite{zhang2018adversarial} use adversarial loss to improve visual quality of estimated sharp images. 
However, their discriminator only considers a single output frame, thus neglecting temporal constraints.
Our training paradigm encourages temporal consistency among multiple output frames, thus regularizing the model to produce temporally-consistent deblurred results.

\noindent{\textbf{Dense correspondence matching.}}~~Many vision tasks can be considered as a correspondence matching problem, \eg~stereo matching~\cite{zhong2018stereo,cheng2020hierarchical} and flow estimation~\cite{zhong2019unsupervised,wang2020displacement}, 
where the essence is the correlation measures for paired images.
%
%
In particular, Nikolaus~\etal~\cite{mayer2016large} directly regress disparity maps from an 1D correlation volume. 
Alex~\etal~\cite{kendall2017end} concatenate features of image frames to form the correlation volume.
The work~\cite{schonberger2018learning} learns to fuse multiple disparity maps to provide a more accurate correlation volume.
%
%
The recent work on optical flow estimation~\cite{teed2020raft} adopts the similar idea but constructs a multi-scale correlation volume.
In their model, coarser correlation volumes are constructed by pooling operations over finer correlation volumes.
%
%
In this way, their models capture distant pixel displacements while preserving fine-grained image details.
%
%
\section{Approach} 
Our model takes as input three consecutive frames from a blurred video and restores the sharp frame in the middle. 
In Fig.~\ref{fig:arch}, we provide an overview of our approach, which consists mainly of three phases: (1) feature extraction; (2) construction of visual correspondence; and (3) sharp frame restoration. All the phases are differentiable and  jointly optimized in an end-to-end trainable architecture.

\subsection{Frame Feature Extraction}
Upon receiving the inputs, our model first applies explicit alignment by warping the neighboring frames towards the reference frame based on the optical flow estimation, as in previous works~\cite{su2017deep,pan2020cascaded}.
In the following, we take the reference frame $\mathbf{I}_i$ and its immediate following frame $\mathbf{I}_{i+1}$ as an example for illustration.
The optical flow estimation module $\mathcal{O}$ takes $\mathbf{I}_i$ and $\mathbf{I}_{i+1}$ and predicts a optical flow field:
\begin{align}
\mathbf{u}_{i, i+1}^x,\,\mathbf{u}_{i, i+1}^y = \mathcal{O}(\mathbf{I}_i, \mathbf{I}_{i+1}),
\end{align}
where $\mathbf{u}_{i, i+1}^x$ and $\mathbf{u}_{i, i+1}^y$ denote the $x,\,y$ components of the estimated flow field, respectively.

The flow field maps each pixel $(x,\,y)$ in ${\mathbf{I}}_{i+1}$ to its corresponding coordinates 
$(x^{\prime},\,y^{\prime})=(x+\mathbf{u}_{i,i+1}^x(x),\,y+\mathbf{u}_{i,i+1}^y(y))$ in ${\mathbf{I}}_{i}$.
We then obtain the aligned neighboring frame $\hat{\mathbf{I}}_{i+1}$ by a bilinear interpolation on $\mathbf{I}_{i+1}$ using the mapped coordinates.
%
We repeat the explicit alignment step for the preceding frame, and acquire the sequence of warped input frames $\{\hat{\mathbf{I}}_{i-1},\,\mathbf{I}_{i},\,\hat{\mathbf{I}}_{i+1}\}$. 
%
%
%

In order to acquire the visual features of aligned frames, we apply an encoder network $\mathcal{F}_\theta$ with trainable parameters $\theta$ to $\{\hat{\mathbf{I}}_{i-1},\,\mathbf{I}_{i},\,\hat{\mathbf{I}}_{i+1}\}$, respectively, and obtain their feature maps with $C$ channels at a reduced resolution.
Namely, the encoder is a mapping $\mathcal{F}_\theta:\mathbb{R}^{H\times W\times 3}\rightarrow\mathbb{R}^{H^{\prime}\times W^{\prime}\times C}$.
In our implementation, the encoder $\mathcal{F}_\theta$ consists of two stride convolutional layers for downsampling the feature maps, and 15 residual blocks, where five are applied at the full resolution, five at the half resolution and five at the quarter resolution.
The output of the frame feature extraction phase is a sequence of frame features $\{\mathcal{F}_\theta(\hat{\mathbf{I}}_{i-1}),\,\mathcal{F}_\theta(\mathbf{I}_{i}),\,\mathcal{F}_\theta(\hat{\mathbf{I}}_{i+1})\}$, where the feature has the dimension $H/4\times W/4\times C$.

\subsection{All-Range Correlation Volume Pyramid}
The explicit alignment method by optical flow builds spatial correspondence between neighboring frames. However, there remain two major drawbacks in such design that might adversely impact the alignment quality.
Firstly, when computing optical flows, the constraint on the velocity smoothness may not be satisfied in the presence of abrupt motions.
Secondly, each blurry image is an integration of multi-frame instance and sharp snapshots with a small motion, but flow fields describe only inter-frame motions while do not capture motions occurring during the exposure time.
Consequently, the optical flow estimation may not be accurate enough to support effective alignment, particularly in dynamic scenes with fast object motions.
%
%
%

Motivated by this observation, we propose a method that is robust to identify visual correspondence with distant pixel displacements in order to mitigate the issue of fast motions. Such a deblurring method is also expected to handle intra-frame motion blurs happening during the exposure.
%
In the following, we present the novel implicit method to construct spatial correspondence by measuring the visual similarity between each frame pair.
%

\noindent\textbf{All-Range Correlation Volume.}~~Convolution kernels operate in a local spatial neighborhood, which limits its ability to capture long-range correspondence. 
Inspired by the recent work in dense correspondence matching~\cite{teed2020raft}, we compute visual correspondence by considering pixel features in all the spatial range at a time.
Given the features of a frame pair, \eg $\mathcal{F}_\theta(\mathbf{I}_{i}),\,\mathcal{F}_\theta(\hat{\mathbf{I}}_{i+1})\in\mathbb{R}^{H^{\prime}\times W^{\prime}\times C}$, a 4D correlation volume $\mathbf{C}(\mathbf{I}_{i},\,{\hat{\mathbf{I}}_{i+1}})\in\mathbb{R}^{H^{\prime}\times W^{\prime}\times H^{\prime}\times W^{\prime}}$ at the full resolution of the feature maps is computed as follows:
\begin{align}
\resizebox{.9\hsize}{!}{$\mathbf{C}(\mathbf{I}_{i},\,\hat{\mathbf{I}}_{i+1})_{xyuv} = \exp\Big(\sum\limits_{c}\mathcal{F}_\theta(\mathbf{I}_{i})_{xyc}\cdot \mathcal{F}_\theta(\hat{\mathbf{I}}_{i+1})_{uvc}\Big)$,\label{eq:cor-vol}}
\end{align}
where $c$ is an index along the channel dimension of frame features.
Eq.~(\ref{eq:cor-vol}) pairs up all the pixel features in the neighboring feature maps, and computes their visual correlation using a radial basis function kernel.
%
This kernel function is introduced to signify the magnitude of strong correlations while suppressing the weak ones.

For each reference frame $\mathbf{I}_i$, we compute two \emph{inter-frame correlation volumes} with the two immediate neighboring frames, \ie~$\mathbf{C}(\mathbf{I}_{i},\,{\hat{\mathbf{I}}_{i-1}})$ and $\mathbf{C}(\mathbf{I}_{i},\,{\hat{\mathbf{I}}_{i+1}})$.
To account for intra-frame motions within the exposure time, we also include an \emph{intra-frame correlation volume} $\mathbf{C}(\mathbf{I}_{i},\mathbf{I}_{i})$, to identify correspondence within the reference frame.
%
%
%
As a result, we acquire three correlation volumes by comparing the reference frame with all the input frames on every pixel pairs across all the spatial range in the feature space.

\noindent\textbf{Correlation Volume Pyramid.}~~In order to enhance the network capacity to learn detailed visual information, we further construct a correlation volume pyramid by incorporating multi-scale features.
%
%
Inspired by the work~\cite{sun2018pwc}, we construct multi-scale representations by pooling over features, instead of input frames, to avoid the heavy computational overhead. 
Specifically, we augment our network with an $L$-layer correlation pyramid $\{\mathbf{C}^k\}_{k=1}^{L}$ by max-pooling over the features of neighboring frames with $L$ kernels of growing sizes.
%
Thereby, the dimension of the correlation volume $\mathbf{C}^{k}$ is $H^{\prime}\times W^{\prime}\times H^{\prime}/{2^k}\times W^{\prime}/{2^k}$.
Different from~\cite{sun2018pwc}, we downsample only the feature maps of the neighboring frame while keeping the resolution of the reference frame features.
As shown in~\cite{teed2020raft}, such design helps to model motions with multiple displacement scales while maintaining high-resolution image details.

\begin{figure}[t!]
\centering
  \includegraphics[width=0.45\textwidth]{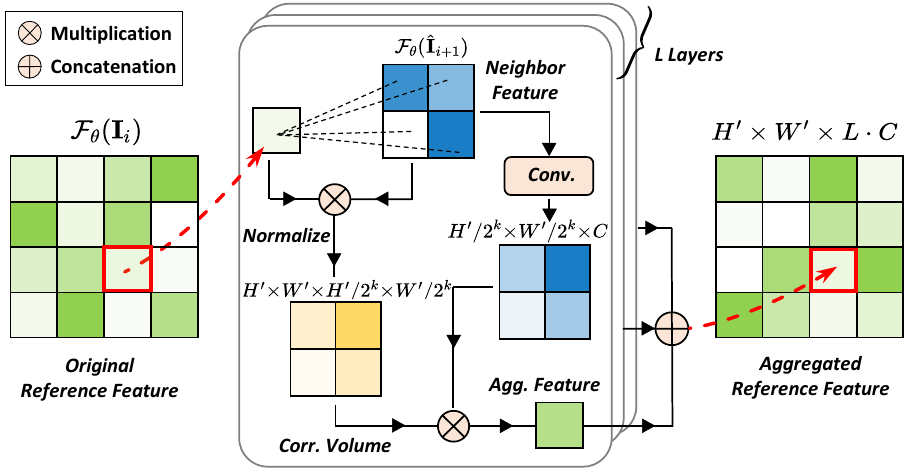}

\caption{Illustration of the correlative aggregation module, which aggregates the features for the reference frame based on the normalized correlation volume.}
\label{fig:lookup}
\end{figure}

\subsection{Deblurring with Volumetric Correspondence}~\label{sec:vol}
In order to restore a sharp reference frame, we develop a correlative aggregation module, which outputs an aggregated feature map of the reference frame by looking up the correlated spatial positions in the neighboring frames. 
%
%
We then concatenate the aggregated feature maps of the reference frame and finally feed them into a reconstruction network to obtain the final deblurring result.

\noindent\textbf{Correlative Aggregation Module.}~~Given a pair of frame features $\{\mathcal{F}_\theta(\mathbf{I}_{i})$, $\mathcal{F}_\theta(\hat{\mathbf{I}}_{i+1})\in\mathbb{R}^{H^{\prime}\times W^{\prime}\times C}\}$,
and the correlation volume pyramid $\{\mathbf{C}^k(\mathbf{I}_i,\,\hat{\mathbf{I}}_{i+1})\}$, where $\mathbf{C}^{k}$ has the dimension $\mathbb{R}^{H\times W\times H/2^k\times W/2^k}$, the correlative aggregation module outputs an aggregated feature map $\rho(\mathbf{I}_{i\leftarrow i+1})$ at the pyramid scale $k$. The aggregated feature map $\rho(\mathbf{I}_{i\leftarrow i+1})$ has the same dimension as the reference frame feature $\mathcal{F}_{\theta}(\mathbf{I}_i)$.
 

We note that the last two dimensions of the correlation volume $\mathbf{C}^k(\mathbf{I}_{i}, \hat{\mathbf{I}}_{i+1})_{xy}\in\mathbb{R}^{H^{\prime}/{2^k}\times W^{\prime}/{2^k}}$ represents a correlation matrix, which describes the correspondence  between the particular feature $\mathcal{F}_{\theta}(\mathbf{I}_i)_{xy}$ and the downsampled neighboring feature map in all the spatial positions.
In order to aggregate the features based on the correspondence, we first normalize the correlation volume along these two dimensions by their sum.
Namely, we compute the normalized correlation volume $\Tilde{\mathbf{C}}^k(\mathbf{I}_i,\,\hat{\mathbf{I}}_{i+1})$ as follows,
\begin{align}
    \Tilde{\mathbf{C}}^k(\mathbf{I}_i,\,\hat{\mathbf{I}}_{i+1})_{xyuv} = \frac{\mathbf{C}^k(\mathbf{I}_i,\,\hat{\mathbf{I}}_{i+1})_{xyuv}}{\sum\limits_u\sum\limits_v{\mathbf{C}}^k(\mathbf{I}_i,\,\hat{\mathbf{I}}_{i+1})_{xyuv}}.
\end{align}



%
Since the correlation volume only provides information on the spatial correspondence and does not consider channel-wise information, we apply another convolution operation $\phi(\cdot)$ to the neighbor feature map.
In this way, we selectively aggregate the channel information and obtain a refined feature map $\phi(\hat{\mathbf{I}}_{i+1})_k\in\mathbb{R}^{H/{2^k}\times W/{2^k}\times C}$.

Next, we aggregate the pixels on the  downsampled feature map based on the correlations.
%
%
Specifically, we compute the aggregated features $\rho(\mathbf{I}_{i\leftarrow i+1})\in\mathbb{R}^{H\times W\times LC}$ as follows:
\begin{align}
    \rho(\mathbf{I}_{i\leftarrow i+1})_{k}&=\Tilde{\mathbf{C}}^k(\mathbf{I}_i,\,\hat{\mathbf{I}}_{i+1})\phi(\hat{\mathbf{I}}_{i+1})_k,\\
    \rho(\mathbf{I}_{i\leftarrow i+1})&={\lVert_{k=1}^{L}}~\rho(\mathbf{I}_{i\leftarrow i+1})_{k},
\end{align}
where the $\lVert$
operator concatenates aggregated feature maps from each volume pyramid layer along the channel dimension.
We also repeat the correlative aggregation once on the preceding neighboring frame and once on the reference frame itself as a self-correlative aggregation. In this way, we obtain aggregated feature maps $\rho(\mathbf{I}_{i\leftarrow i-1})$ and $\rho(\mathbf{I}_{i\leftarrow i})$, respectively.

We concatenate all the three aggregated feature maps of the reference frame along the feature dimension, each of dimension $H\times W\times L\cdot C$, thus obtaining the final feature map.
Lastly, we follow an reconstruction network to restore the sharp reference frame. We implement the reconstruction network as in~\cite{tao2018scale} considering its effectiveness. Detailed architecture will be provided in the supplementary materials.

\subsection{Progressive Generative Adversarial Learning}\label{tab:training}
Inspired by~\cite{pan2020cascaded}, we split the training procedure into stages and enable the model to progressively achieve fine-detailed restoration.
To further elaborate, we use $T_i^s$ to denote a single training step in $s$-th stage to restore $\mathbf{I}_{i}$. For the first stage, $T^1_i$ takes $\mathbf{I}_{i-1}$, $\mathbf{I}_{i}$ and $\mathbf{I}_{i+1}$ and outputs $\mathbf{R}_{i}^1$ as a restoration for $\mathbf{I}_{i}$.
The later stages take the output of the previous stage as input. 
For example, $T_i^2$ takes  $\mathbf{R}_{i-1}^1$, $\mathbf{R}_{i}^1$ and  $\mathbf{R}_{i+1}^1$ in order to restore $\mathbf{I}_{i}$.
In this way, a two-staged learning requires five consecutive frames as input.  

Adversarial training succeeds in generating and restoring visually realistic images~\cite{ledig2017photo,karras2017progressive,kupyn2018deblurgan}.
On video deblurring, Zhang~\cite{zhang2018adversarial} uses a discriminator to distinguish between the restored single frame and its ground-truth, which does not consider temporal constraints.
In contrast, we exploit the proposed progressive training scheme and design an adversarial loss to encourage models to output temporally-consistent results. 
In this regard, we propose a temporal discriminator $\mathcal{D}_{\zeta}$,  parameterized by $\zeta$, that distinguishes $\vec{\mathbf{R}}_i=\{\mathbf{R}_{i-1}^1\,\mathbf{R}_{i}^2,\,\mathbf{R}_{i+1}^2\}$ from their ground-truth sharp images $\vec{\mathbf{S}}_i=\{\mathbf{S}_{i-1},\,\mathbf{S}_{i},\,\mathbf{S}_{i+1}\}$.
In specific, $\mathcal{D}_\zeta$ consists of four 3D convolution layers followed by a global pooling to extract temporal features for the ground-truth and restored frame sequences. It then predicts whether the sequence is a ground-truth sequence or not using a binary classifier. 
%
The adversarial loss is defined as follows,
\begin{align}
    L_{adv}=\mathbb{E}_\xi[\log(D_\zeta(\vec{\mathbf{S}}))] + \mathbb{E}_\chi[\log(1-D_\zeta(\vec{\mathbf{R}}))],
\end{align}
where $\xi$ is a distribution of the ground-truth sharp sequences and $\chi$ is that of the input sequences. The discriminator $\mathcal{D}_\zeta$ and the deblurring model are optimized adversarially in the typical minimax fashion.   In this way, we encourage the model to produce a consistent sequence that visually resemble the ground-truth.
We combine the proposed adversarial loss with $L_1$ loss for reconstruction, \ie $L_{total} = L_1 + \alpha L_{adv}$, where $\alpha=0.1$.
\section{Experiments}
In this section, we first present experimental results by evaluating our proposed model ARVo 
on standard datasets. 
Then we conduct ablation studies to investigate the effects of each component in the proposed model.

\subsection{Implementation Details}
We implement ARVo in PyTorch~\cite{paszke2019pytorch}. 
We adopt a three-layer correlation pyramid to achieve a trade-off between video deblurring quality and computational efficiency.
We use PWC-Net~\cite{sun2018pwc} for optical flow estimation considering its robustness in real-life domains.
During the training, we feed five consecutive frame patches of 256$\times$256 resolution to the model and optimize it using Adam optimizer~\cite{kingma2014adam}. We set the initial learning rate as 10$^{-4}$ and reduce it by half every 200 epochs. We also use random flipping and rotations to augment the training dataset.
The weights are shared across all the progressive stages for a compact model.

\subsection{Datasets and Evaluation Metrics}
We evaluate the proposed methods on two datasets DVD~\cite{su2017deep} and a newly introduced HFR-DVD dataset.
We adopt peak signal-to-noise ratio (PSNR) and structural similarity (SSIM)~\cite{wang2004image} for evaluating the deblurring results.

\noindent\textbf{DVD~\cite{su2017deep}.}~This dataset has 71 videos with 6,708 blurry-sharp pairs, where 61 videos are used for training (5,708 pairs) and 10 videos (1,000 pairs) for testing.
These videos are captured at 240 fps with mobile phones and DSLR.
The DVD dataset is widely adopted in the community and has promoted the development of video deblurring models.

\noindent\textbf{HFR-DVD.}~~To simulate realistic blurs, one typical approach is capturing videos at a high frame rate and then averaging multiple consecutive sharp frames for blurs.
In this way, the quality of the synthetic blurry images is affected by the exposure time of devices. In particular, a lower frame rate leads to longer idling time between adjacent exposures, yielding unnatural spikes in the blur trajectories.
In this regard, previous datasets mainly tackle the issue by introducing virtual frames.
For example, Su~\etal~\cite{su2017deep} generate inter-frame images based on optical flow estimation.
Nah~\etal~\cite{nah2019ntire} proposed to increase the frame rate of sharp videos by interpolations with a pretrained neural network.
In contrast, we construct a High-Frame-Rate Dataset for Video Deblurring (HFR-DVD) by capturing videos at 1,000 fps with a SONY DSC-RX10 IV camera, thus avoiding the artifacts caused by automatic frame interpolation.

\begin{figure}[t!]
\centering
  \includegraphics[width=0.48\textwidth]{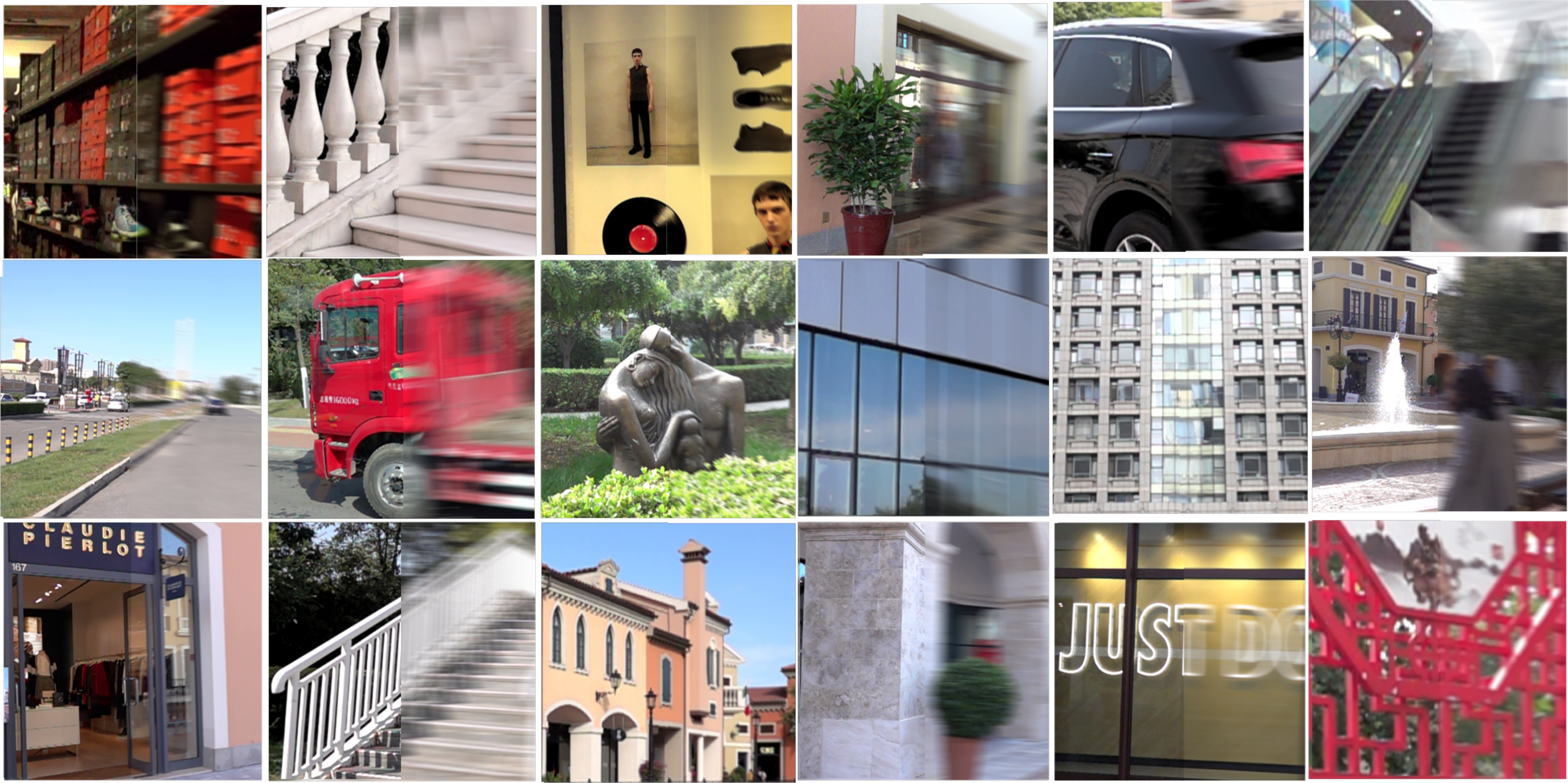}

    \caption{Example frames from our newly introduced HFR-DVD dataset, showing a diversity in scenes and motions. The left half of the frame is sharp while the right half shows the synthetic blurs.}
\label{fig:dataset}
\end{figure}

When capturing videos, we pay attentions to the video quality and the diversity of scenes. A few example frames are shown in Fig.~\ref{fig:dataset}.
We synthesize the blurs by averaging $41$ neighboring sharp frames. To reduce the noises, we downsize each frame to 960$\times$540 resolution.
We then subsample the high-frame-rate video temporally by keeping the middle frame every $N$ frames. We choose $N$ randomly from the interval [38, 44]. In this way, we finally obtain blurry videos at 25 fps.
%
In total, we acquire 120 videos for training and 30 videos for testing, each with 90 frames, amounting to 13,500 blurry-sharp pairs altogether.
We will release the HFR-DVD dataset publicly soon.

\subsection{Comparisons with the State-of-The-Art} \noindent\textbf{Synthetic Blurry Videos.}~We first compare our model ARVo with previous deblurring methods~\cite{hyun2015generalized,su2017deep,tao2018scale,zhang2018adversarial,wang2019edvr,zhou2019spatio,xiang2020deep,pan2020cascaded} on the DVD dataset. Among them, the work~\cite{hyun2015generalized} is based on variational models whereas the rest works rely on deep neural networks.
%

\begin{table*}[t!]
\caption{Quantitative comparisons on the DVD dataset~\cite{su2017deep}. Following~\cite{pan2020cascaded}, we report PSNR and SSIM metrics considering all the restored video frames. As can be seen, our model performs favorably against previous approaches in both metrics. }\label{table:dataset}
\centering


\resizebox{\textwidth}{!}{
\begin{tabular}{cccccccccc}
\toprule
              Methods    & Kim~\etal~\cite{hyun2015generalized}   &     EDVR~\cite{wang2019edvr}
         & Tao~\etal~\cite{tao2018scale}        & Su~\etal~\cite{su2017deep}         & DBLRNet~\cite{zhang2018adversarial} & STFAN~\cite{zhou2019spatio} & Xiang~\etal~\cite{xiang2020deep} & TSP~\cite{pan2020cascaded} & \textbf{ARVo (Ours)}
         \\\midrule
PSNR &    26.94  & 28.51 &    29.98 & 30.01 & 30.08 & 31.15 & 31.68 & 32.13 & \textbf{32.80} \\
SSIM  & 0.8158  & 0.8637 & 0.8842 & 0.8877 & 0.8845 & 0.9049 & 0.9157 & 0.9268 & \textbf{0.9352}   
\\
\bottomrule
\end{tabular}
}~\label{tab:dvd}
\end{table*}

\begin{figure*}[t!]
\centering

\setlength\tabcolsep{1.5pt}
\begin{tabular}{cccccc}
\includegraphics[width=0.16\textwidth]{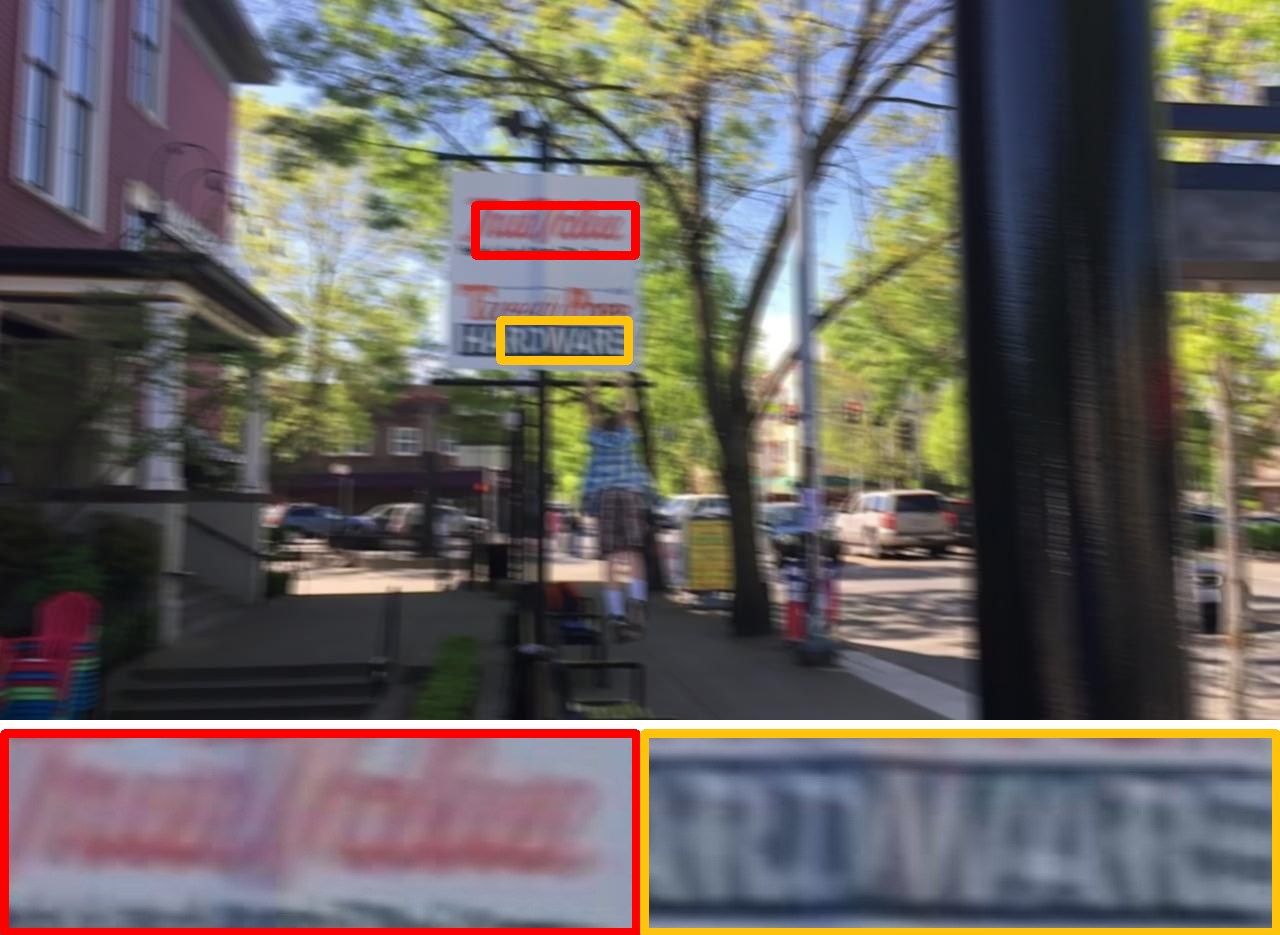}&
\includegraphics[width=0.16\textwidth]{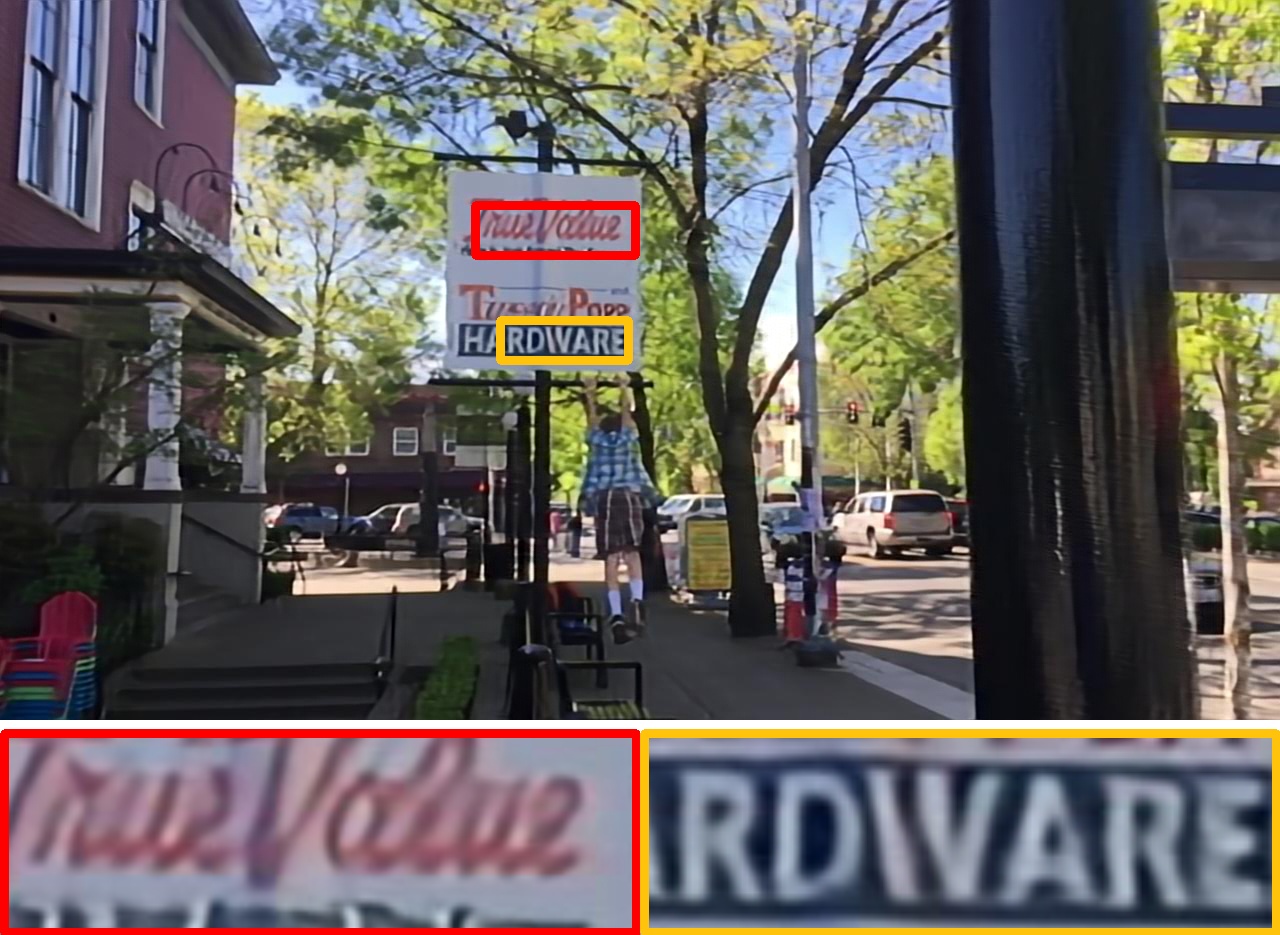}&
\includegraphics[width=0.16\textwidth]{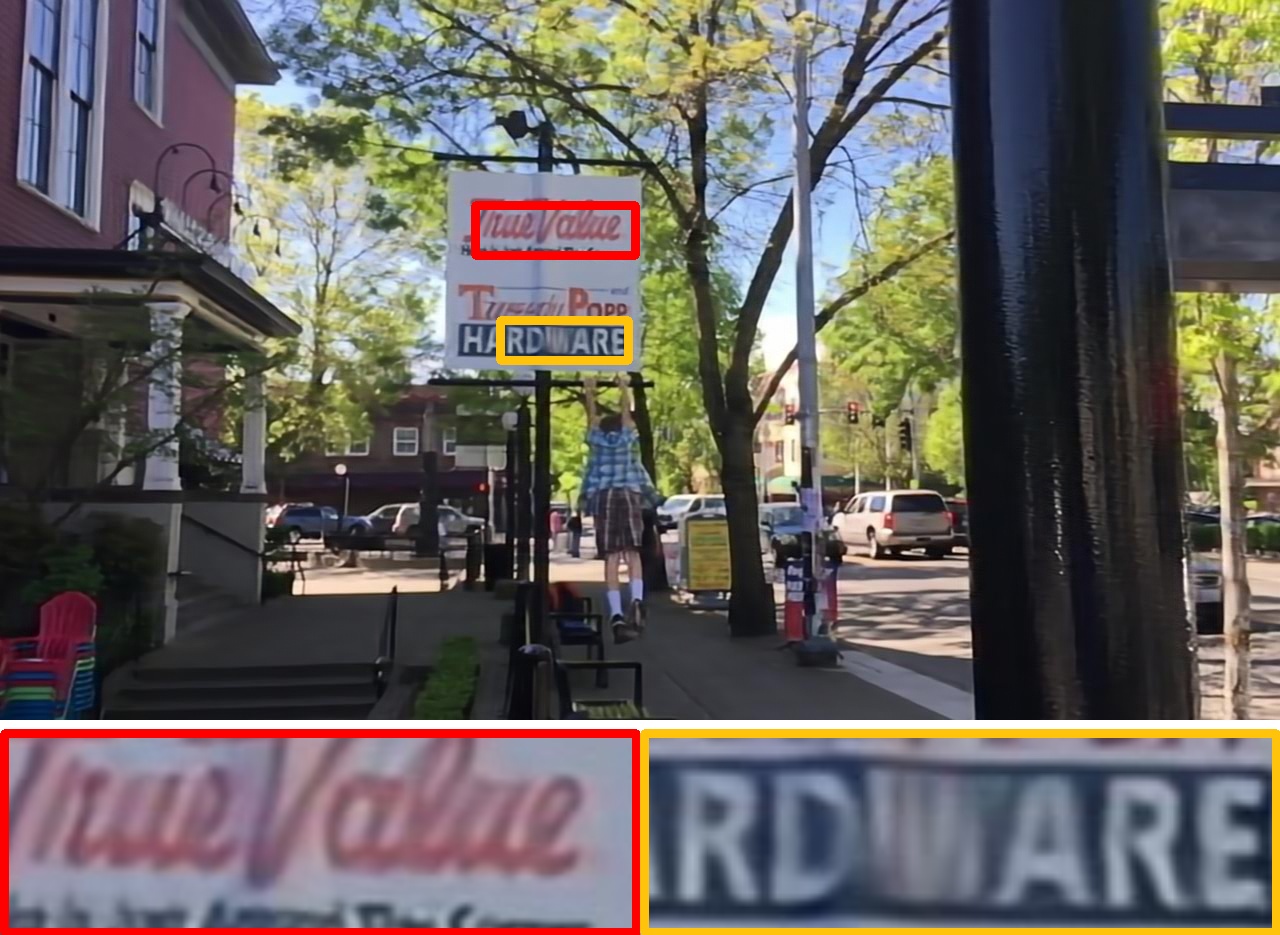}&
\includegraphics[width=0.16\textwidth]{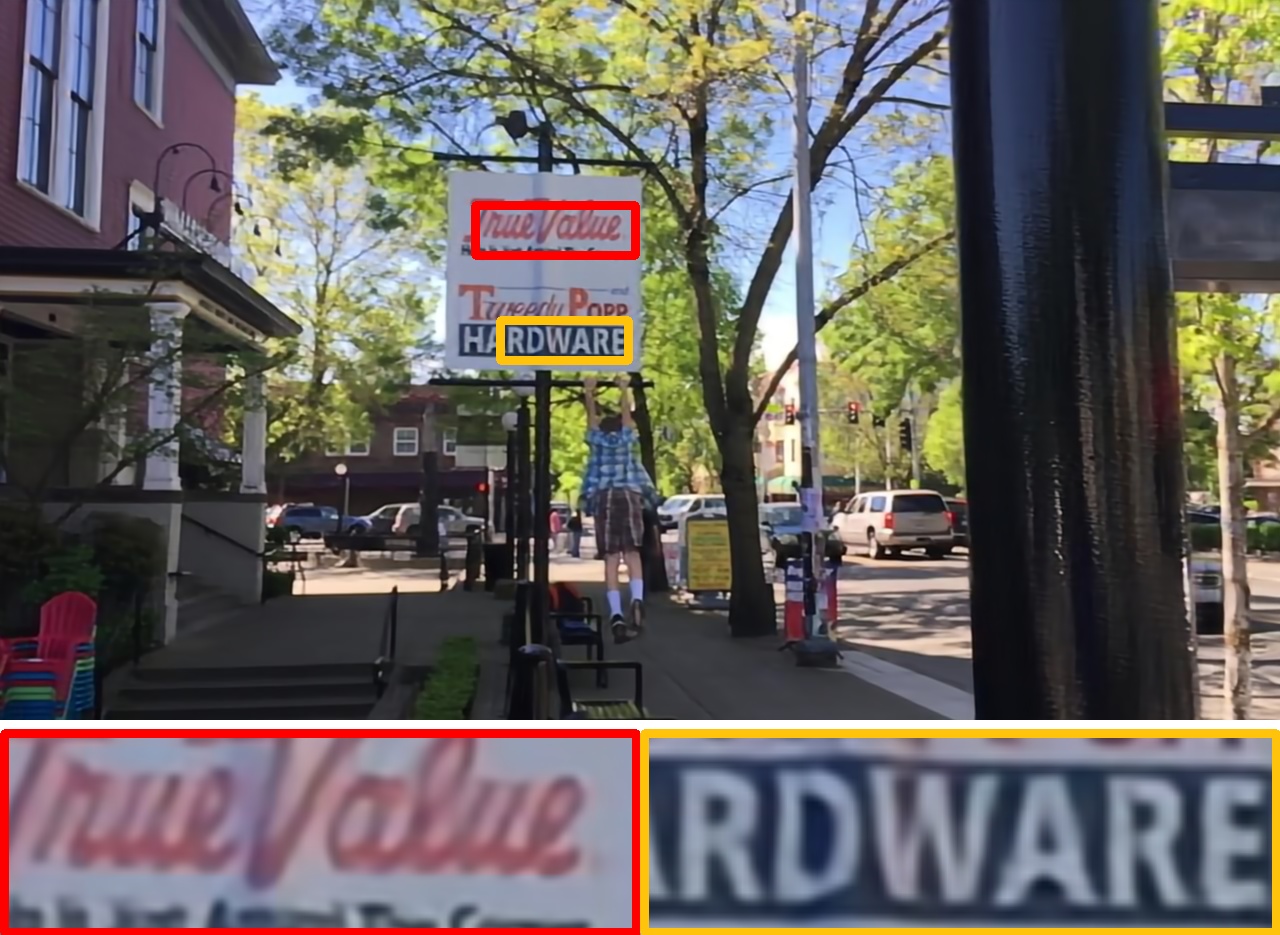}&
\includegraphics[width=0.16\textwidth]{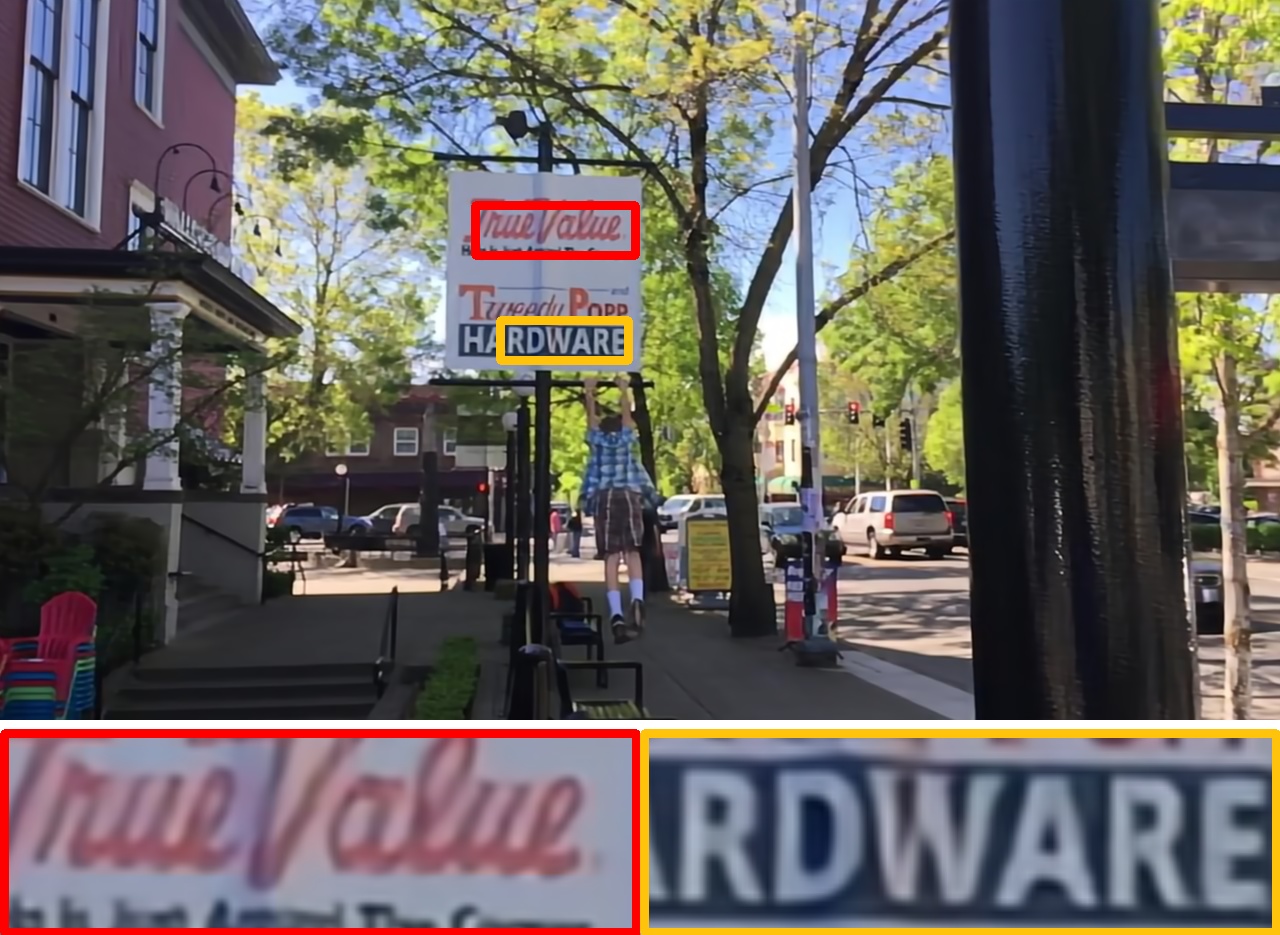}&
\includegraphics[width=0.16\textwidth]{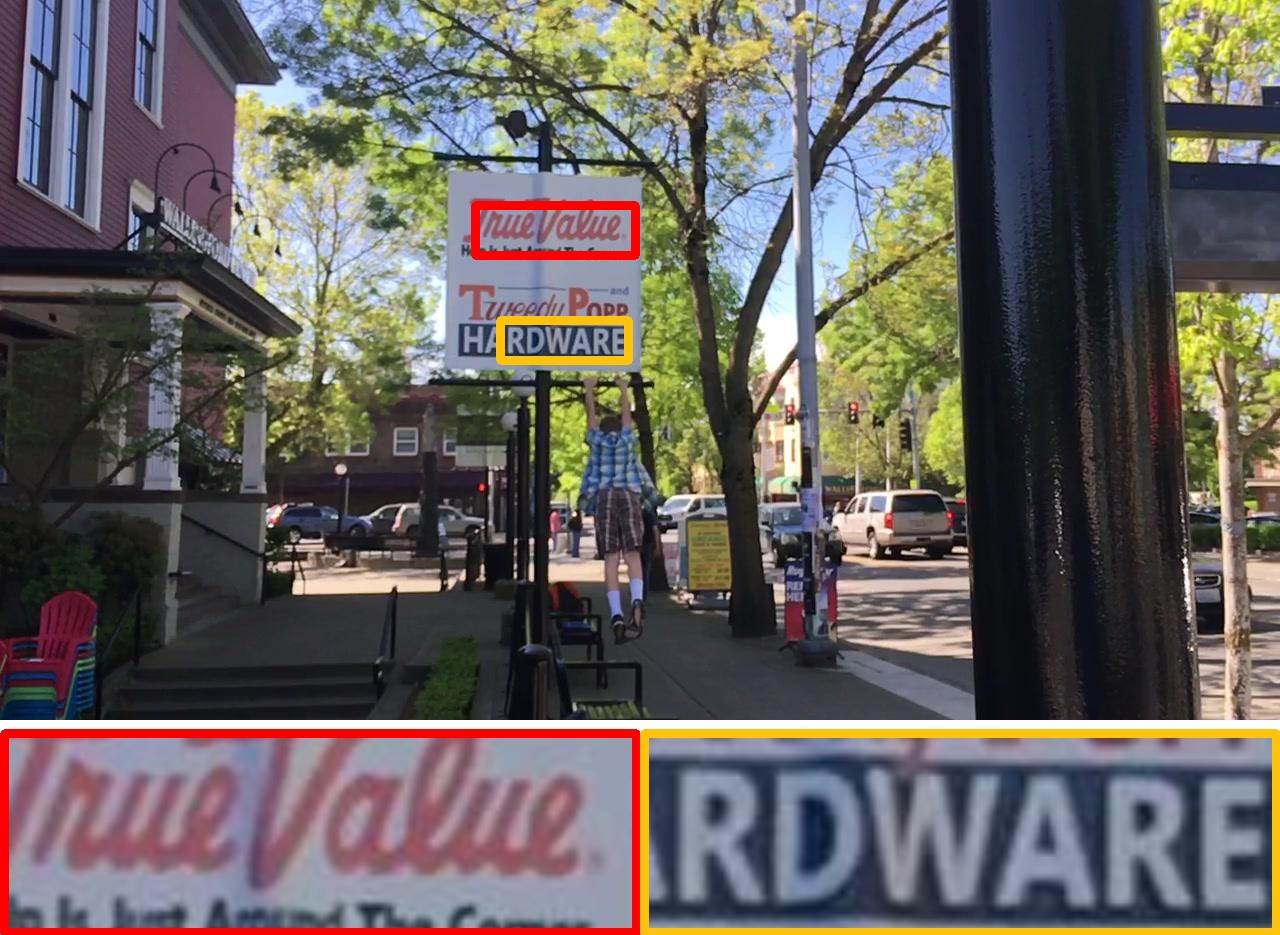}\\

\includegraphics[width=0.16\textwidth]{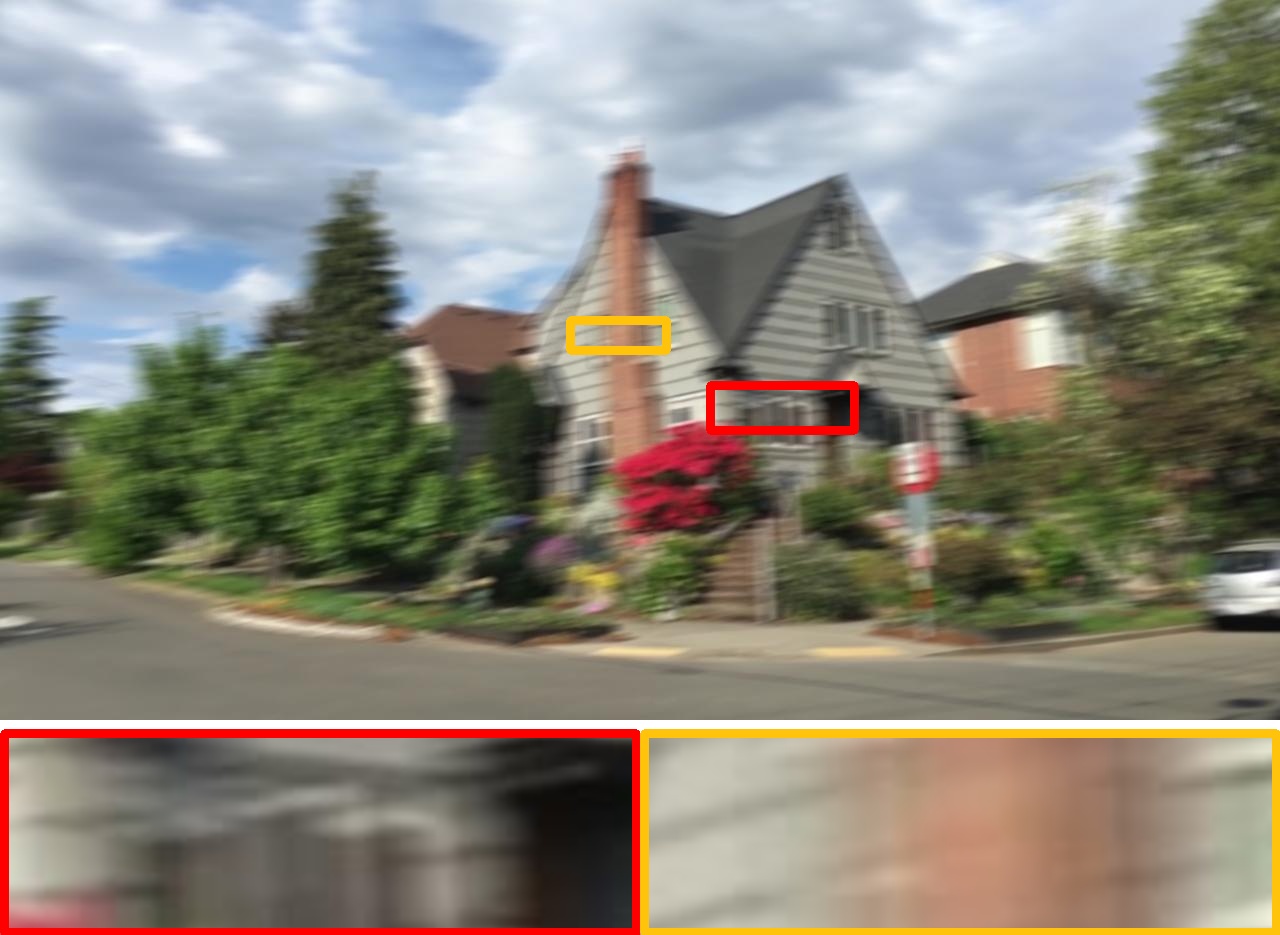}&
\includegraphics[width=0.16\textwidth]{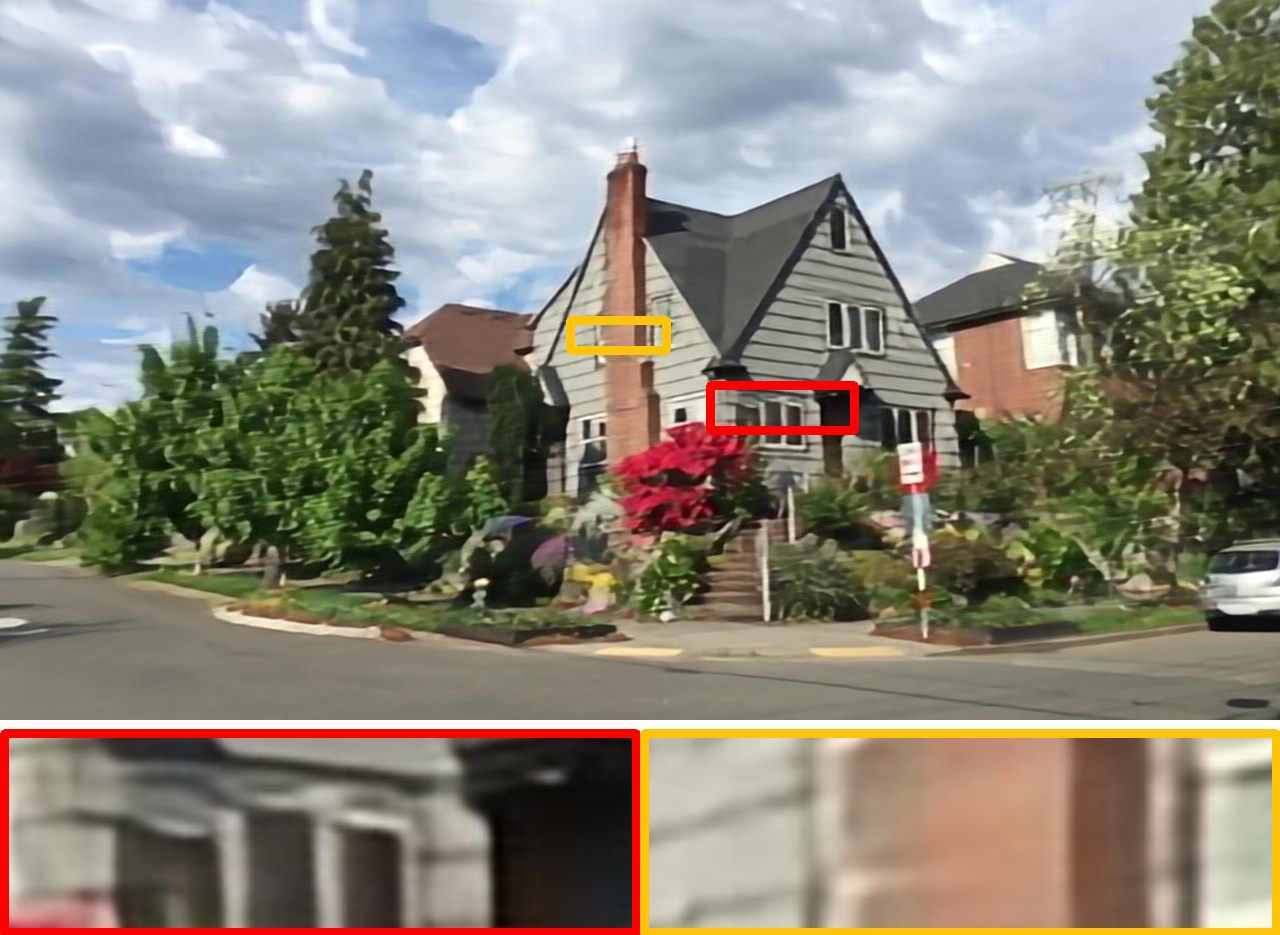}&
\includegraphics[width=0.16\textwidth]{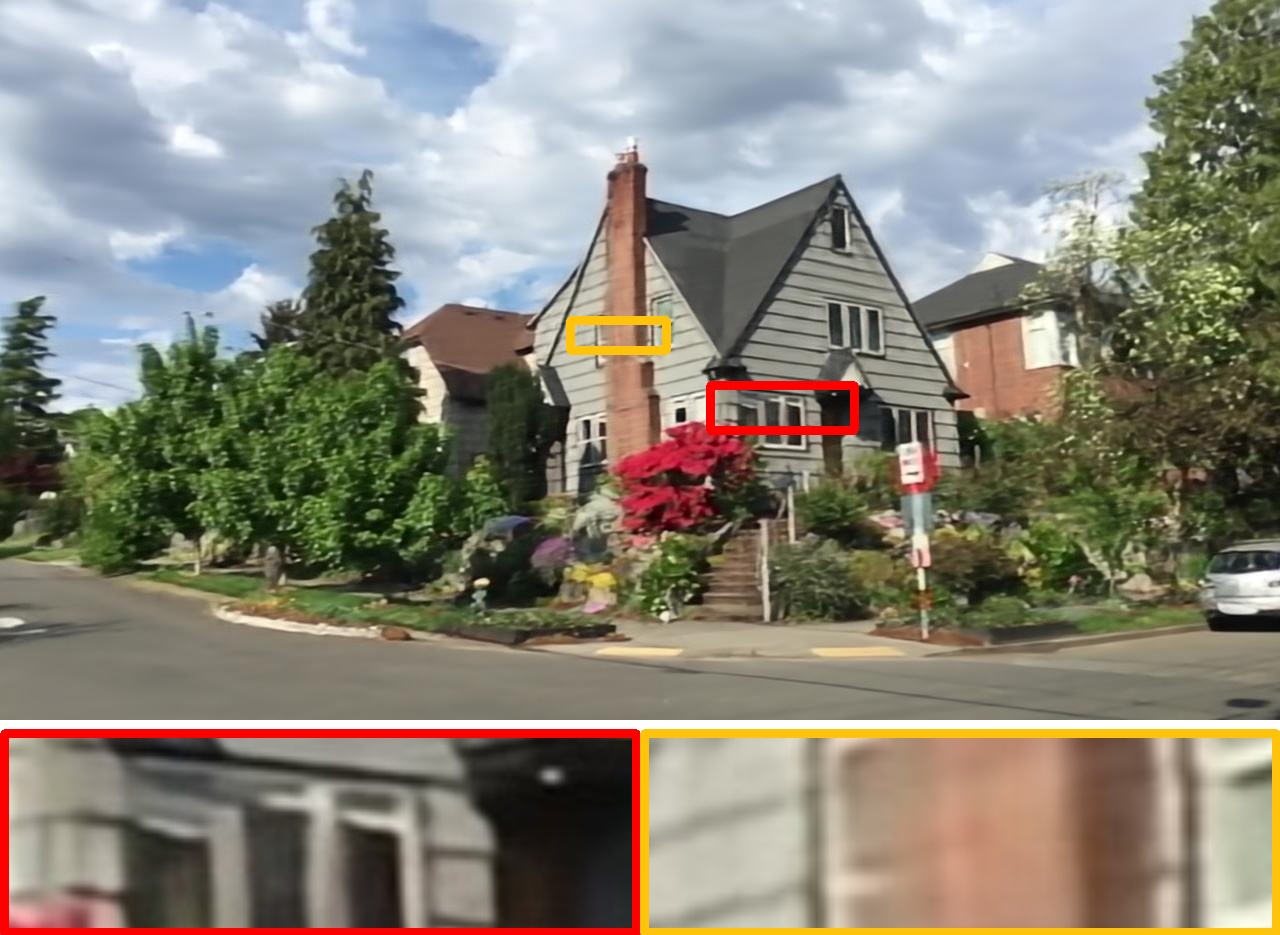}&
\includegraphics[width=0.16\textwidth]{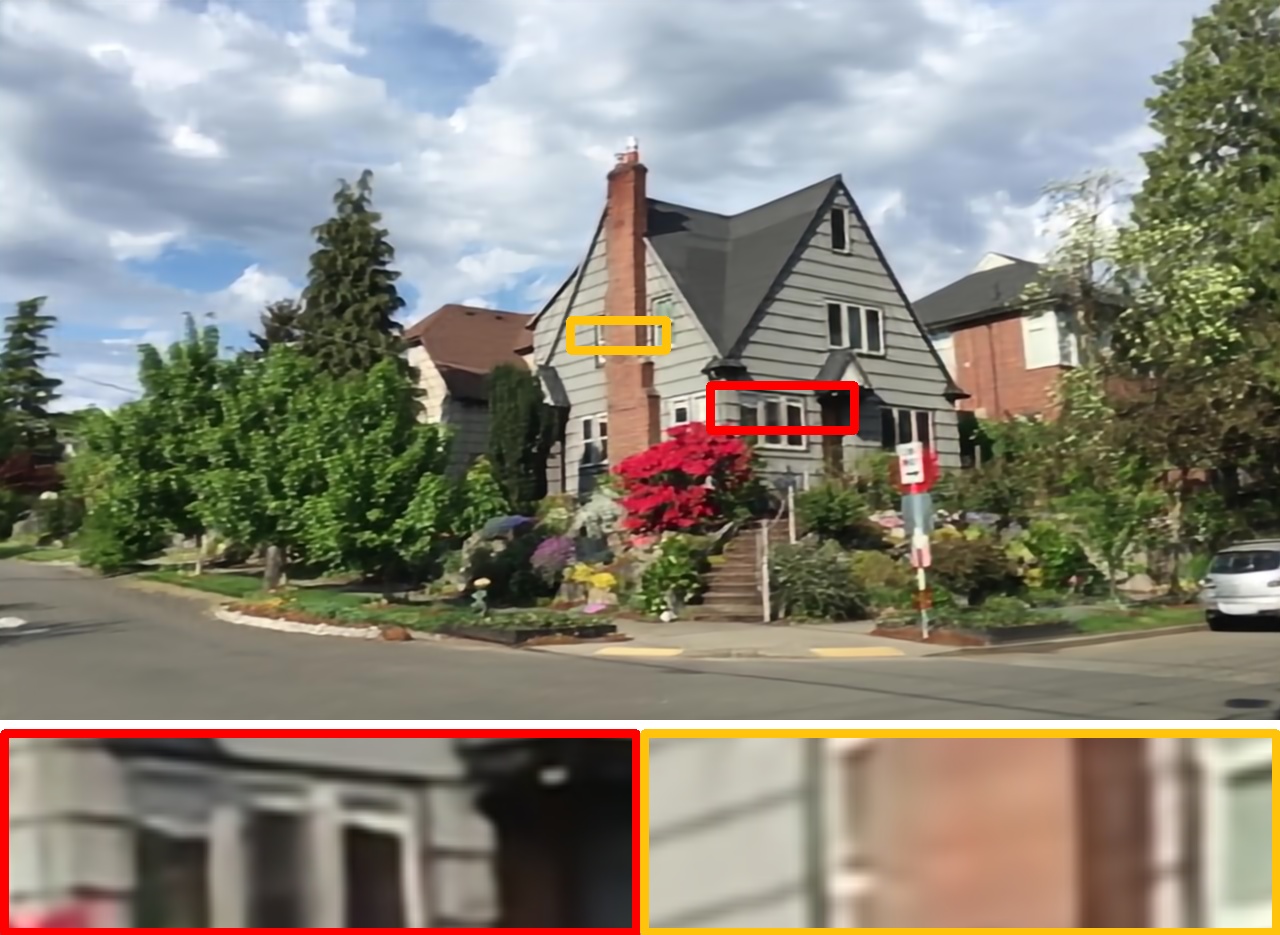}&
\includegraphics[width=0.16\textwidth]{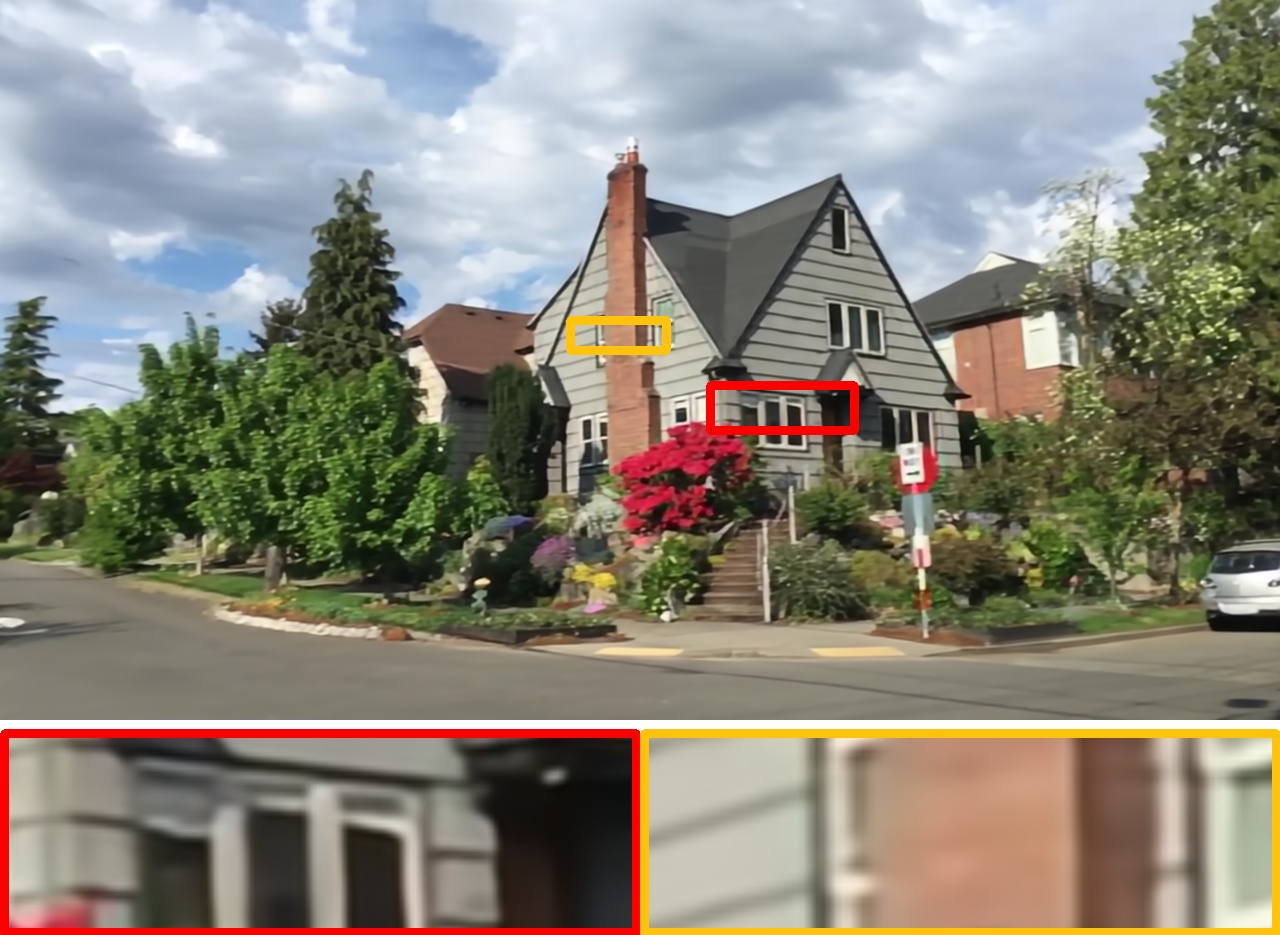}&
\includegraphics[width=0.16\textwidth]{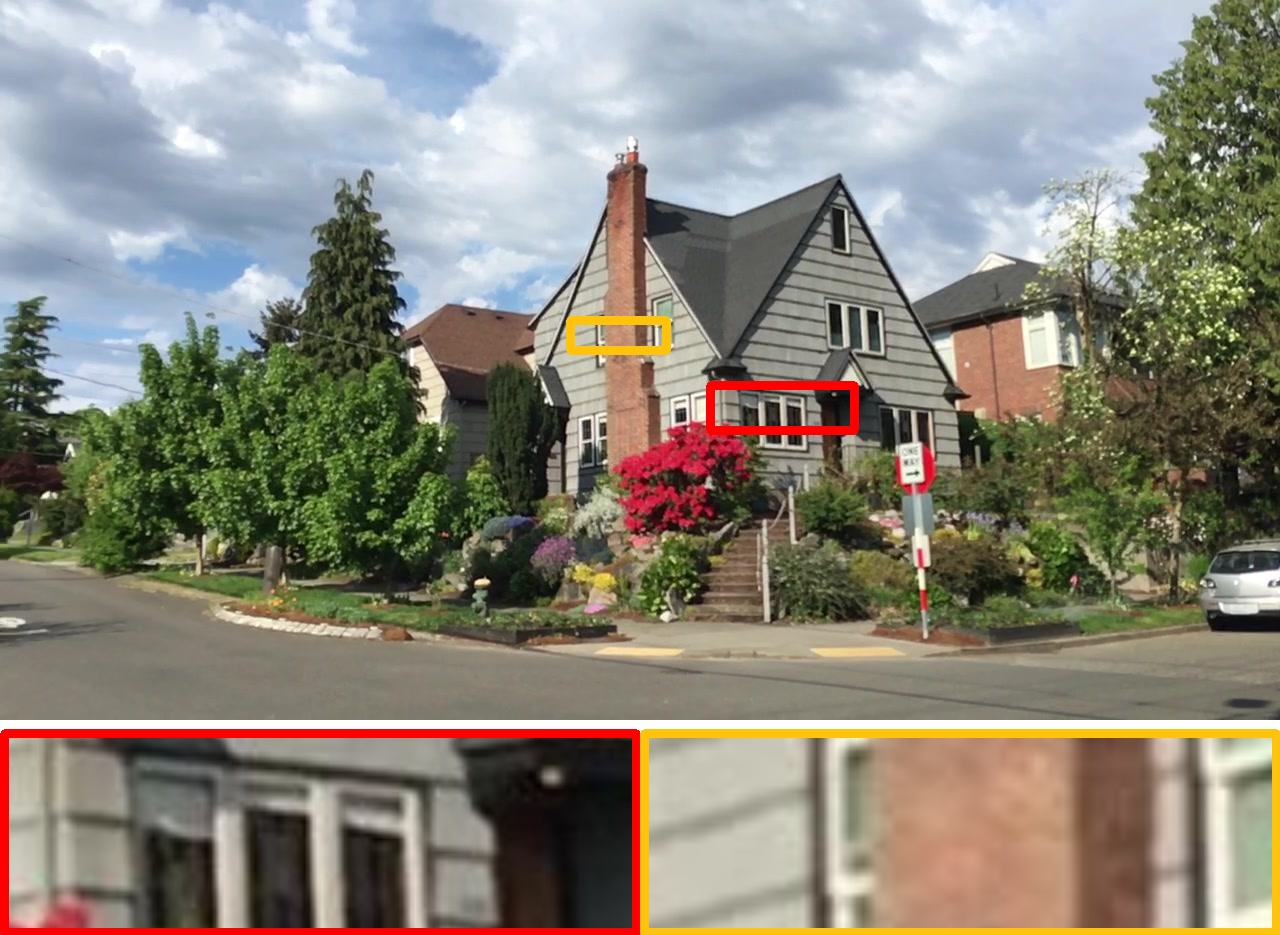}\\

(a) Input & (b) Su~\etal~\cite{su2017deep} & (c) STFAN~\cite{zhou2019spatio} & (d) TSP~\cite{pan2020cascaded} & (e) ARVo (Ours) & (f) Ground-truth

\end{tabular}


\caption{Qualitative comparisons on the DVD~\cite{su2017deep} dataset. Our model benefits from constructing pixel pair correspondence in the feature space, thus being able to better utilize neighboring sharp patches to improve the restoration quality.}
\label{fig:dvd}
\end{figure*}

Table~\ref{tab:dvd} shows the quantitative comparisons on the DVD dataset. As shown, our model performs favorably against previous works in both metrics, indicating a better restoration quality.
In Fig.~\ref{fig:dvd} we show the deblurred images output from different models.
The work~\cite{su2017deep} designs an end-to-end convolution models for deblurring video frames. However, they rely mainly on explicit alignment methods such as homography or optical flows only,  therefore are less favorable to restore abrupt motion blurs. 
The method~\cite{zhou2019spatio} uses only two consecutive frames as the contextual information, thus do not fully utilize the temporal information in the video. As can be seen in Figure~\ref{fig:dvd}, it fails to fully restore the structure of scene objects.
We further note that the method~\cite{pan2020cascaded} computes a temporal sharpness prior to guide the model to focus on areas with severe blurs. However, the prior only considers a small range of pixels, thereby, it is less effective when fast motions are present.
In contrast, our model takes in multiple consecutive frames as input, providing stronger video contextual clues.
By implicitly constructing the pixel-pair correspondence across all the spatial range, our model is able to effectively captures pixel correlations with large displacements. 
This is particularly effective for blurs due to fast motions.
As a result, our proposed model restores better the detailed scene content.  

We further evaluate our method quantitatively on the newly introduced HFR-DVD dataset.
We remark that the HFR-DVD dataset exhibits more fast motions compared to DVD dataset, which further challenges the capacity of video deblurring models. 
We select five models~\cite{su2017deep,zhang2018adversarial,zhou2019spatio,pan2020cascaded} among the best performers based on the quantitative evaluations on DVD as the baseline methods.
%

As Table~\ref{tab:hfrdvd} shows, our model performs favorably against baseline methods in both metrics. 
Also, we notice that it outperforms the previous best model by a slightly larger margin than on the DVD dataset. %
This validates our motivation for tackling fast motion blurs with all-range spatial correspondence. 
In Fig.~\ref{fig:hfr}, we show qualitatively the deblurring results on the HFR-DVD dataset.
Due to the inter-frame correlation volume computed on all the pixel pairs, our model utilizes better the video context thus producing sharper restored frames with finer visual details.

\begin{figure*}[t!]
\centering

\setlength\tabcolsep{1.5pt}
\begin{tabular}{cccccc}
\includegraphics[width=0.16\textwidth]{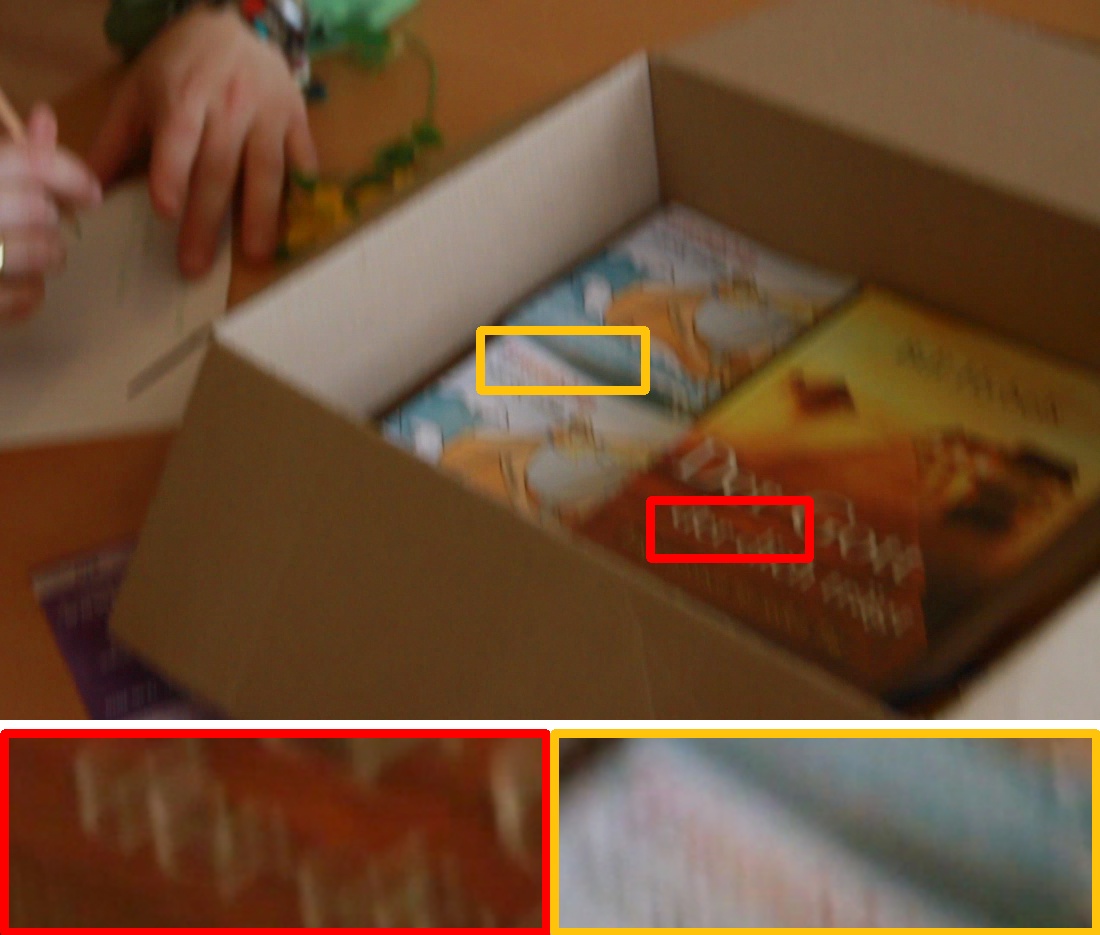}&
\includegraphics[width=0.16\textwidth]{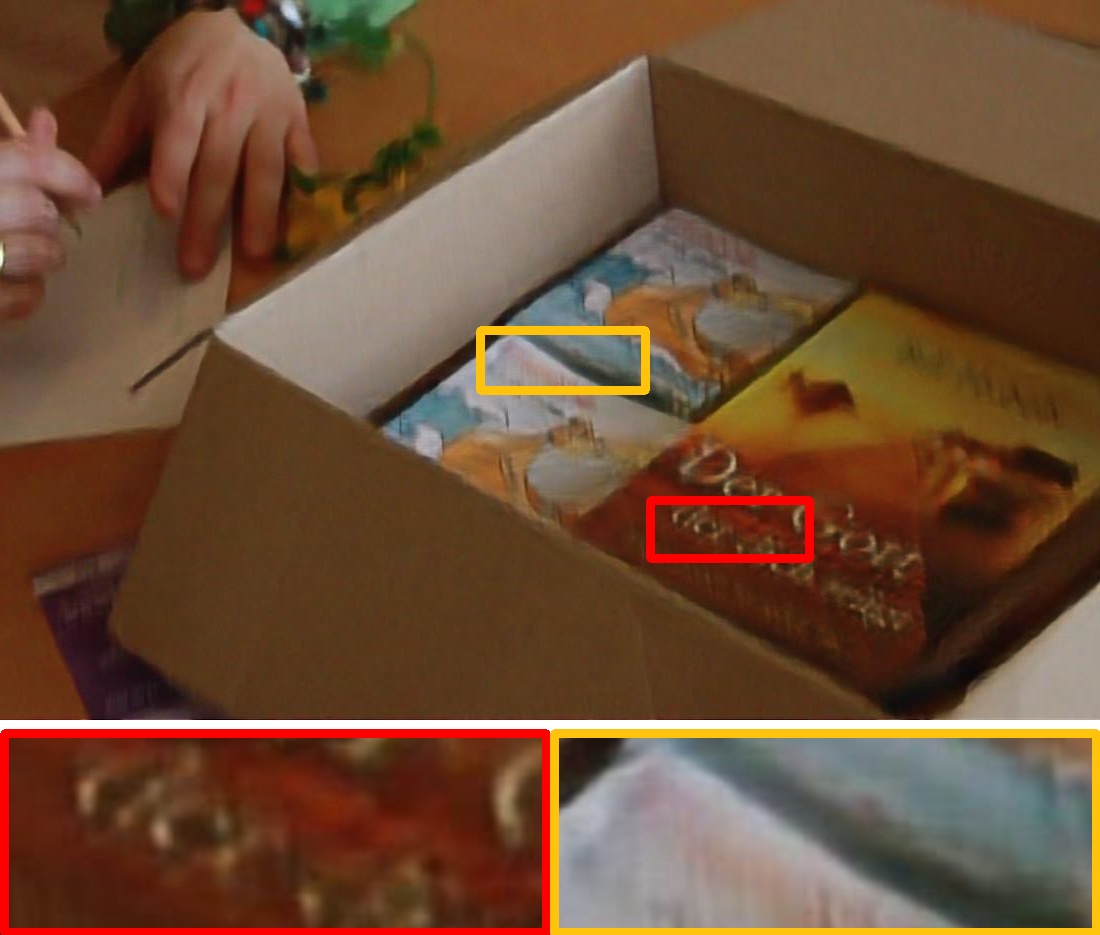}&
\includegraphics[width=0.16\textwidth]{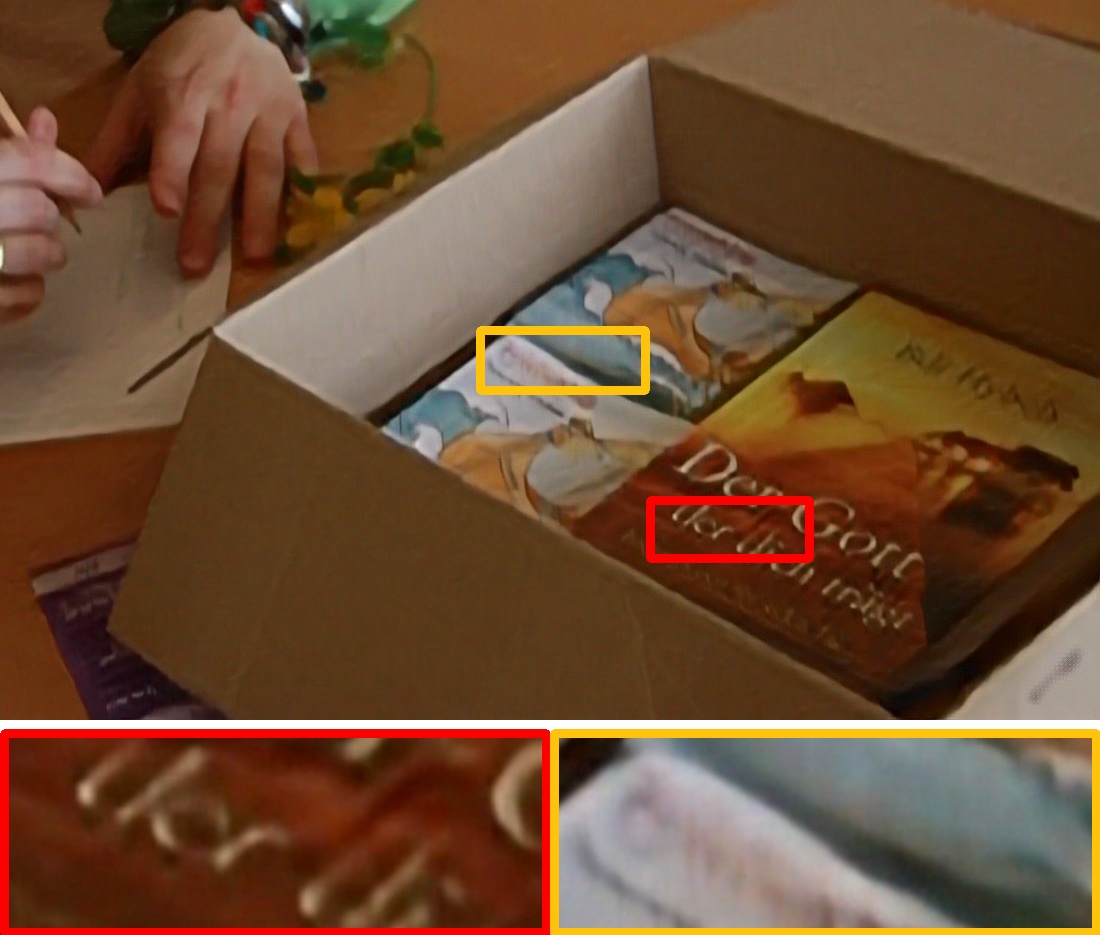}&
\includegraphics[width=0.16\textwidth]{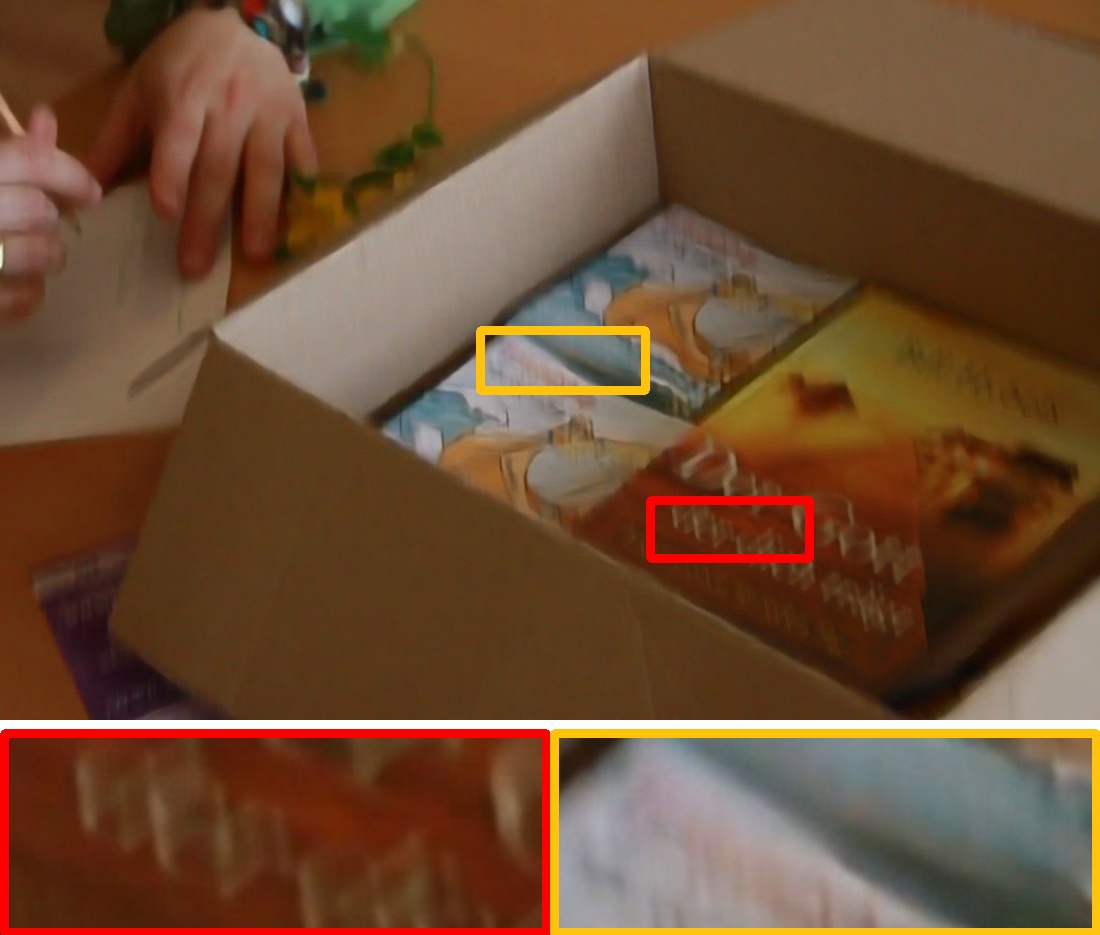}&
\includegraphics[width=0.16\textwidth]{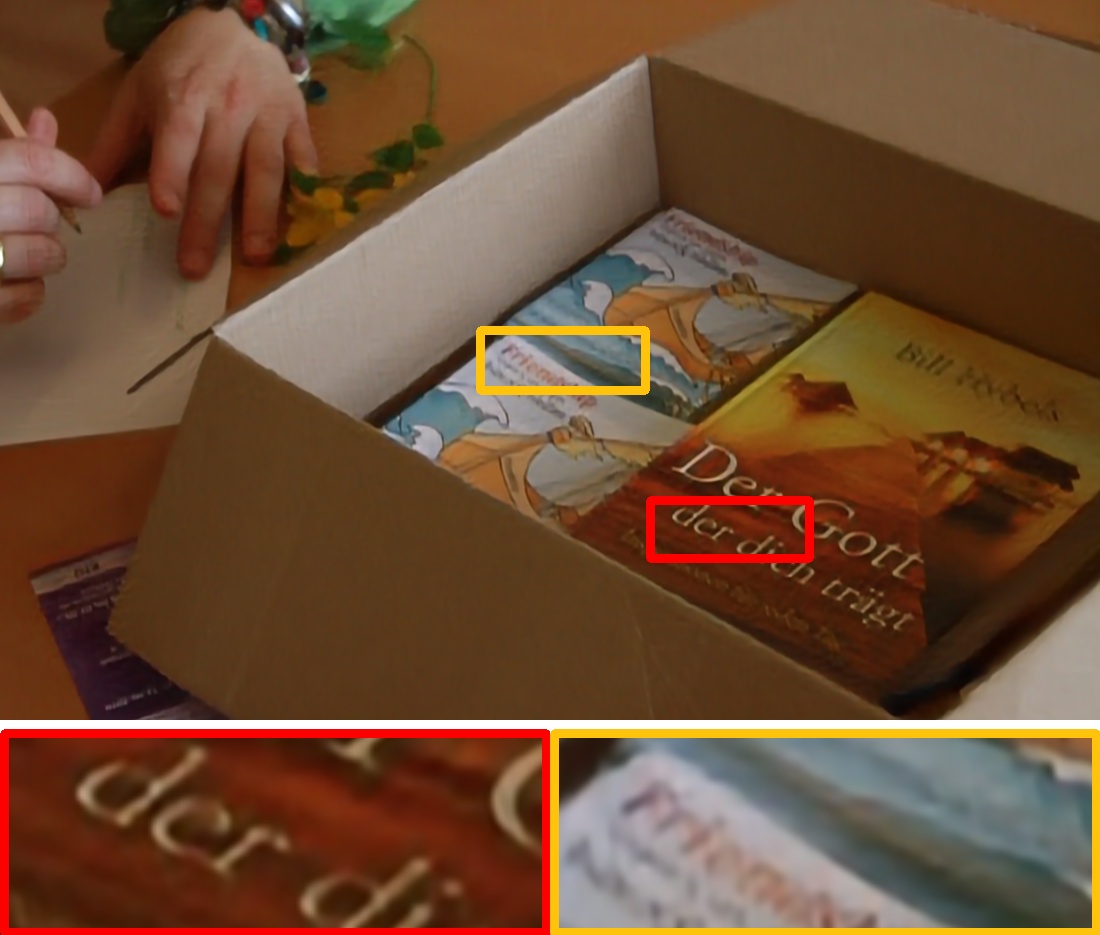}&
\includegraphics[width=0.16\textwidth]{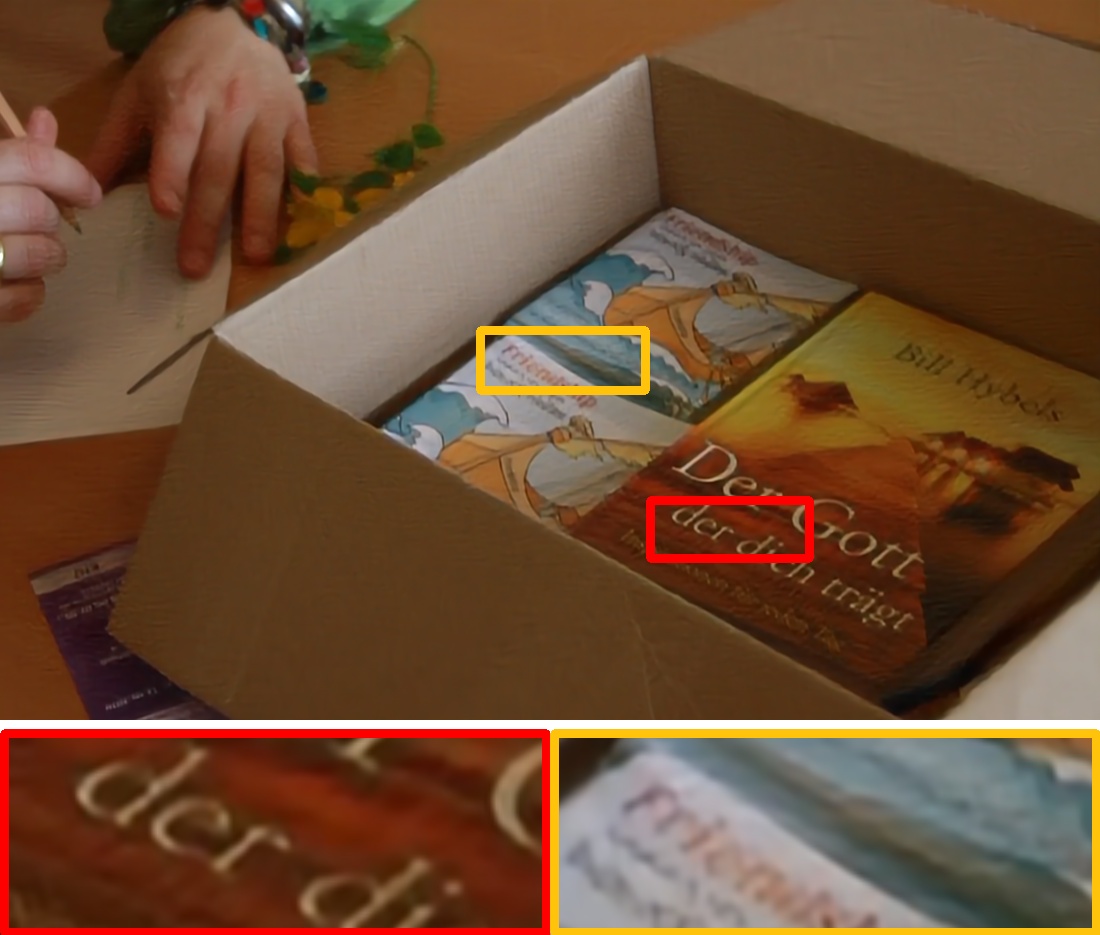}\\
(a) Input & (b) Su~\etal~\cite{su2017deep} & (c) DBLRNet~\cite{zhang2018adversarial} & (d) STFAN~\cite{zhou2019spatio} & (e) TSP~\cite{pan2020cascaded} & (f) ARVo (Ours)

\end{tabular}


\caption{Qualitative comparisons on the real blurry frames. As shown, our model provides sharper restored frames while preserving more fine-grained visual details, which demonstrates the effectiveness of our model in real-life scenarios.}
\label{fig:real}
\end{figure*}
\noindent\textbf{Real Blurry Videos.}
In order to validate the generalization ability of our proposed model, we validate ARVo on the real-world blurry videos from~\cite{su2017deep}. As can be seen from Fig.~\ref{fig:real}, our model restores better structural information than the previous methods, which demonstrates a promising video deblurring outcome in real-life scenarios.
\begin{table}[t!]
\caption{Quantitative comparisons using the HFR-DVD dataset. We adhere to the same evaluation protocol as in Table~\ref{tab:dvd} for a fair and consistent comparison. }\label{table:dataset}
\centering


\resizebox{0.45\textwidth}{!}{
\begin{tabular}{cccccc}
\toprule
        Methods &\cite{su2017deep}         &\cite{zhang2018adversarial} &\cite{zhou2019spatio} &\cite{pan2020cascaded} &\textbf{Ours}
         \\\midrule
PSNR & 27.73 & 27.79 & 28.48 & 29.71 & \textbf{31.15} \\
SSIM  &  0.8512 & 0.8505 & 0.8560 & 0.8822 & \textbf{0.9063}  
\\
\bottomrule
\end{tabular}
}~\label{tab:hfrdvd}
\end{table}
\subsection{Model Analysis and Discussions}
In this section, we take DVD dataset as an example to analyze the effects of multiple components in our proposed method.
We also investigate how the proposed correlative aggregation module affects the feature learning to better understand the behavior of our models.

\noindent\textbf{Effect of Correlation Pyramid}~~We propose to construct a correlation volume pyramid in the feature space in order to handle large pixel displacements. 
To analyze the effect of correlation volumes, we experiment with different choices of pyramid layers with $L=0, 1,2,3,4$.
When $L=0$, we do not construct volume pyramid.
Instead, we directly feed the frame features from the encoder to the reconstruction network. As shown in Table~\ref{tab:vol-layers}, learning features using correlation volumes evidently improves the deblurring results even with only one layer of pyramid.
By adding pyramid layers at multiple spatial scales, we observe further improvement in the restoration quality. 
This is because the increased receptive fields help to capture better visual context and long-range motion dependencies.
We notice no significant benefit by building more than three layers of pyramid. This shows that a pyramid layer with a downsampling rate of 16 is too coarse to provide useful information to enhance the deblurring model.

\noindent\textbf{Effect of Alignment Methods}~~While typical optical flow-based alignment methods provide useful spatial alignment, it is less effective to restore sharp images with fast motions. We complement the typical explicit alignment method by constructing correlation volumes in the feature space. 
In order to understand the contribution of different alignment methods, we conduct experiments to use (i) no alignment (no align); (ii) only optical flow-based alignment (O.F.); (iii) only correlation volumes (Vol.). 
Results are shown in Table~\ref{tab:align}. As can be seen, using no alignment method leads to less favorable results. This is because alignment methods, either explicit or implicit, help the model to identify pixel correspondence more easily. 
We also remark that the optical flow-based methods and correlation volumes are complementary in their alignment effects. Though using correlation volumes solely provides slightly better results than explicit alignment, our model provides the best restoration quality when combining both alignment methods.


\noindent\textbf{Effect of Progressive Training}~~By training the model progressively, we allow the network to consider wider video contexts and gradually restore the sharp reference frame. 
In Table~\ref{tab:prog}, we show the influence of progressively training the deblurring model. As shown, our model gains an improvement by training for two stages while benefiting less from considering further frames. This is because the scenes change largely as the input video segment gets longer, thus making it harder to find related sharp patches. We decide on two stages to achieve a trade-off between deblurring efficiency and quality. 
We further remark that our model surpasses previous methods even trained with a single stage, again showing its effectiveness.
\begin{table}[t]
\caption{Effects of using different layers of correlation pyramids on the video deblurring result. We highlight the setting in our final model in bold.}\label{table:ablate-vol}
\centering
\resizebox{0.45\textwidth}{!}{
\begin{tabular}{cccccc}
\toprule
        Num. Layers &0         &1 &2 &3 &4
         \\\midrule
PSNR & 31.85 & 32.45 & 32.69 & \textbf{32.80} & 32.81 \\
SSIM  &  0.9132 & 0.9310 & 0.9335 & \textbf{0.9352} & 0.9352
\\
\bottomrule
\end{tabular}
}~\label{tab:vol-layers}
\end{table}
\begin{table}[t]
\caption{Effects of using different alignment methods. The O.F. column shows results with only optical flow-based alignment; the Vol. column shows results when using only correlation volumes; the O.F. + Vol. shows results when using both methods together. The final setting is in bold.}\label{table:dataset}
\centering
\resizebox{0.45\textwidth}{!}{
\begin{tabular}{ccccc}
\toprule
        Align Methods          & no align & O.F. & Vol. & O.F. + Vol.
         \\\midrule
PSNR & 31.19 & 31.85 & 32.01 & \textbf{32.80}\\
SSIM  & 0.9055  & 0.9132 & 0.9192 & \textbf{0.9352}
\\
\bottomrule
\end{tabular}}~\label{tab:align}
\end{table}
\begin{table}[t]
\caption{Effects of progressive training using different number of stages and input frames. The final setting is in bold.}
\centering

\resizebox{0.4\textwidth}{!}{\begin{tabular}{cccc}
\toprule
        Num. Stages (Frames) &1 (3)         &2 (5) &3 (7)
         \\\midrule
PSNR & 32.28 & \textbf{32.80} & 32.83 \\
SSIM  &  0.9254 & \textbf{0.9352} & 0.9349
\\
\bottomrule
\end{tabular}~\label{tab:prog}}
\end{table}

\begin{figure}[t!]
\centering
  \includegraphics[width=0.48\textwidth]{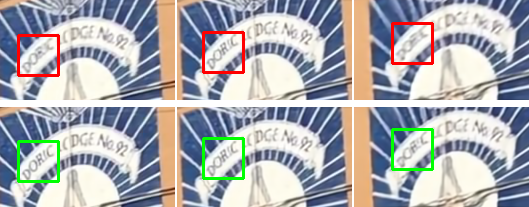}

    \caption{The middle image is from stage 2 while the rest are from stage 1. \textbf{Top}: without adversarial loss.
    \textbf{Bottom}: with adversarial loss. Adversarial training helps to maintain the visual quality in consecutive output
    frames.}
\label{fig:gans}
\end{figure}

\noindent\textbf{Effect of Generative Adversarial Training}~~By introducing generative adversarial training, we do not observe a significant impact on the evaluation metrics, obtaining 32.79 (-0.01) for PSNR and 0.9354 (+0.02) for SSIM. This might attribute to that the adversarial loss does not optimize towards the pixel-values in regard to the ground-truth. However, we do observe a perceivable difference on the restored images by using the adversarial loss. 
In Fig.~\ref{fig:gans}, we demonstrate the effect of using adversarial loss for training. As can be seen, the generative adversarial training helps to maintain the visual consistency across frames.

\begin{figure}[t!]
\centering

\setlength\tabcolsep{1.5pt}
\begin{tabular}{cc}
\includegraphics[width=0.23\textwidth]{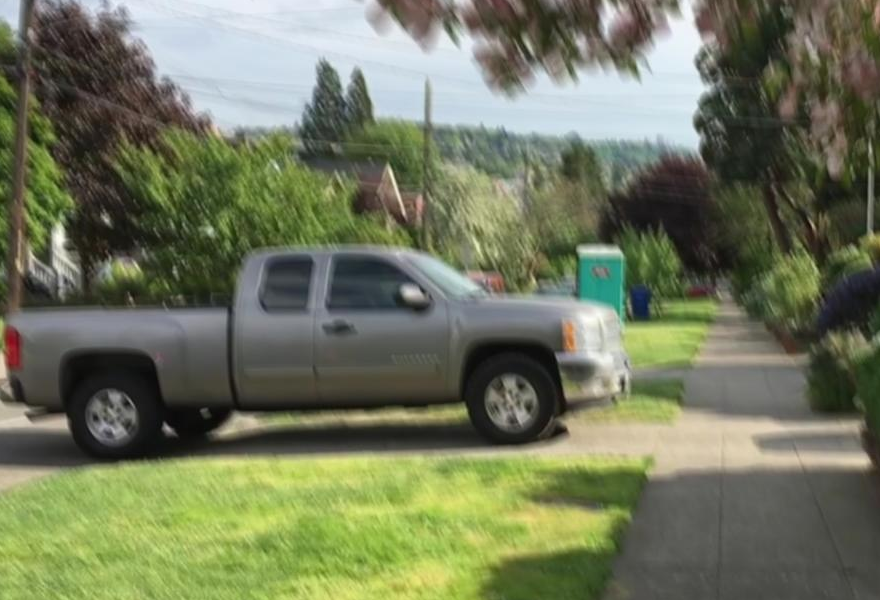}&\includegraphics[width=0.23\textwidth]{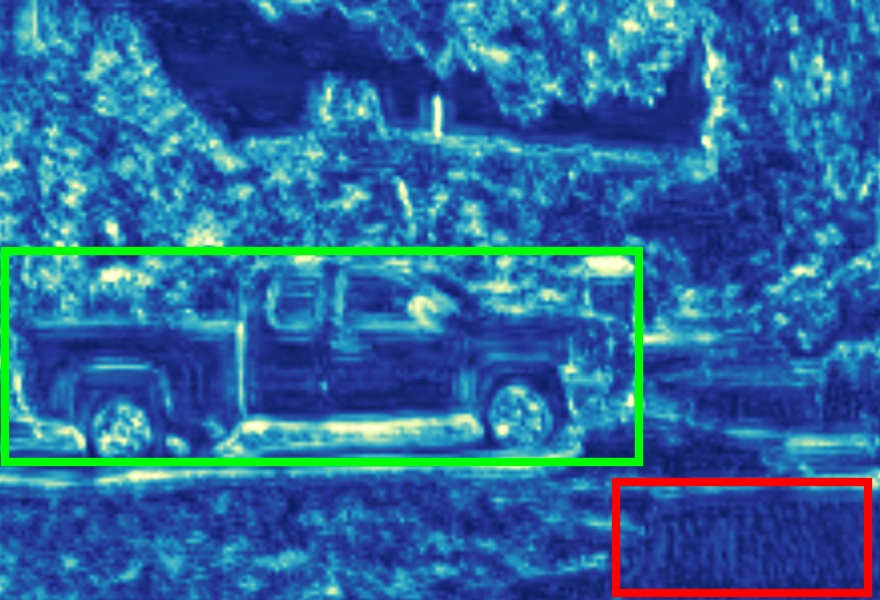}
\end{tabular}
\caption{\textbf{Left}: reference frame.
\textbf{Right}: Visualization of the feature update after applying the correlative aggregation. 
Lighter positions indicate more significant changes.
The green box shows a blurry patch that is updated evidently. The red box shows that the network updates less on a shaded road patch with less blurs. }\label{fig:vis}
\end{figure}

\noindent\textbf{Feature Visualization for Correlative Aggregation}
In order to gain further insights into the proposed model, we compare the feature of the reference frame before and after the correlative aggregation and show the changes on the feature map in Fig.~\ref{fig:vis}. We observe that the network effectively learns to refine areas with severe blurs, while applying less updates on patches with less blurs.
%




\section{Conclusion}
In this work, we propose a novel implicit method to construct spatial correspondence for video deblurring. The method builds a correlation volume pyramid by matching pixel-pairs between the reference frame and the neighboring frame in all the spatial range. 
Based on such a correlation volume pyramid, we develop a correlative aggregation module to enhance the features of the reference frame.
Sharp frames are then restored from the enhanced feature maps via a reconstruction network.
We design a generative adversarial training scheme to optimize the model progressively.
We also present a newly-collected high-frame-rate dataset for video deblurring (HFR-DVD), featuring sharper frames and more realistic blurs.
Our model achieves favorable performance on both datasets quantitatively and qualitatively, establishing a new state-of-the-art for video deblurring. 

{\small
\bibliographystyle{ieee_fullname}
\bibliography{egbib}
}

\end{document}